\documentclass[11pt]{article}

\usepackage[final]{acl}

\usepackage{times}
\usepackage{latexsym}

\usepackage[T1]{fontenc}

\usepackage[utf8]{inputenc}

\usepackage{microtype}

\usepackage{inconsolata}

\usepackage{graphicx}

\usepackage{subcaption}
\usepackage{enumitem}
\usepackage{svg}
\usepackage[svgnames]{xcolor}
\usepackage{booktabs}

\usepackage{colortbl}
\usepackage{pgfmath}
\usepackage[most]{tcolorbox} 
\usepackage{siunitx}         
\usepackage{pgf}             
\usepackage{ifthen}          

\usepackage{amssymb}

\usepackage{booktabs}
\usepackage{multirow}
\tcbset{
  myboxstyle/.style={
    colback=gray!10!white,   
    colframe=gray!70!black,  
    fonttitle=\bfseries,     
    coltitle=white,          
    boxrule=0.3mm,           
    arc=2mm,                 
    left=1mm, right=1mm,     
    top=1mm, bottom=1mm      
  }
}

\newcommand{\perc}[2]{%
  \pgfmathsetmacro{\increase}{((#2 - #1) / #1 * 100)}%
  \pgfmathparse{round(\increase)}%
  \pgfmathtruncatemacro{\increaseInt}{\pgfmathresult}%
  \pgfmathtruncatemacro{\displayPerc}{abs(\increaseInt)}%
  \pgfmathsetmacro{\colorlevel}{max(0, min(100, \displayPerc * 3))}%
  \ifdim \increase pt > 0pt
    \def\arrowSymbol{$\uparrow$}%
    \def\cellColor{green!50!white}%
  \else
    \def\arrowSymbol{$\downarrow$}%
    \def\cellColor{red!50!white}%
  \fi

  \edef\temp{\noexpand\cellcolor{\cellColor!\colorlevel} \num[round-precision=0]{#2}\, \arrowSymbol\displayPerc\%\ }%
  \temp%
}

\usepackage{soul,xcolor}
\newcommand{\redhl}[1]{\sethlcolor{red!15}\hl{#1}}
\newcommand{\bluehl}[1]{\sethlcolor{blue!15}\hl{#1}}
\newcommand{\brownehl}[1]{\sethlcolor{brown!15}\hl{#1}}
\newcommand{\purehl}[1]{\sethlcolor{blue!20}\hl{#1}}
\title{Can AI-Generated Persuasion Be Detected?\\Persuaficial Benchmark and AI vs. Human Linguistic Differences}

\author{
 \textbf{Arkadiusz Modzelewski\textsuperscript{1,2,3}}, 
 \textbf{Pawe{\l} Golik},
 \textbf{Anna Ko{\l}os\textsuperscript{1}},
 \textbf{Giovanni Da San Martino\textsuperscript{2}}
\\
\\
 \textsuperscript{1}NASK National Research Institute, Poland \\
 \textsuperscript{2} University of Padua, Italy \\
  \textsuperscript{3} Polish-Japanese Academy of Information Technology, Poland \\
 \small{
   \textbf{Correspondence:} \href{mailto:contact@amodzelewski.com}{contact@amodzelewski.com}
 }
}

\begin{document}
\maketitle
\begin{abstract}
Large Language Models (LLMs) can generate highly persuasive text, raising concerns about their misuse for propaganda, manipulation, and other harmful purposes. This leads us to our central question: \textit{Is LLM-generated persuasion more difficult to automatically detect than human-written persuasion?} To address this, we categorize controllable generation approaches for producing persuasive content with LLMs and introduce Persuaficial, a high-quality multilingual benchmark covering six languages: English, German, Polish, Italian, French and Russian. Using this benchmark, we conduct extensive empirical evaluations comparing human-authored and LLM-generated persuasive texts. We find that although overtly persuasive LLM-generated texts can be easier to detect than human-written ones, subtle LLM-generated persuasion consistently degrades automatic detection performance.
Beyond detection performance, we provide the first comprehensive linguistic analysis contrasting human and LLM-generated persuasive texts, offering insights that may guide the development of more interpretable and robust detection tools.
\end{abstract}

\section{Introduction}
\label{sec:introduction}
Persuasive writing, which uses rhetorical techniques and devices to influence audiences, has become central to modern communication \cite{gass2022persuasion}. We live in an era where artificial intelligence increasingly shapes propaganda and persuasive communication in news, political discourse, and social media \cite{BergmanisKorats2024, goldstein2024persuasive}. Large Language Models, now widely used in writing and communication tasks, demonstrate a growing potential to produce persuasive text and influence public opinion \cite{pauli2025measuring, bai2025llm, breum2024persuasive, karinshak2023working}. Several studies have explored how effectively LLMs can identify persuasive language \cite{sprenkamp2023large, panasyuk2025synthclassify}. Yet, to the best of our knowledge, no prior work has addressed whether the automatic detection of LLM-generated persuasion is more challenging than detecting persuasion in human-written texts. Understanding this distinction is crucial, as it reveals the extent to which current detection systems may be vulnerable to increasingly sophisticated AI-driven persuasive content. Furthermore, although previous research has highlighted the importance of mitigating and defending against AI-generated persuasion \cite{burtell2023artificial, el2024mechanism}, it has largely focused on persuasion detection, leaving the linguistic differences between human-written and LLM-generated persuasive texts unexplored. Understanding these differences could deepen our knowledge of AI-driven persuasion and support the development of more effective automatic detection methods. To address these gaps, we investigate two key research questions: \textbf{RQ1} \textit{Is controllably generated AI persuasion harder for LLMs to detect in a zero-shot setting than human-written persuasion?} and \textbf{RQ2} \textit{What are the linguistic differences between controllable LLM-generated and human-written persuasive texts?}.

To address our first research question, we introduce \textbf{Persuaficial}, a newly constructed benchmark of artificially generated persuasive texts. Persuaficial is a novel multilingual resource comprising synthetic persuasive content produced using four generation approaches inspired by \citet{chencan}. The dataset is created in a controlled manner, leveraging human-written texts drawn from established datasets \cite{piskorski-etal-2023-multilingual, tan2016winning, moral2023overview}. In our experiments, we evaluate the detectability of AI and human-written persuasive texts using four different LLMs, including commercial closed models and open-weight models. Our analysis focuses on English, but we also provide analysis on five additional languages: German, French, Italian, Polish, and Russian.

To address RQ2, we conducted an analysis using the StyloMetrix tool\footnote{\url{https://github.com/ZILiAT-NASK/StyloMetrix}}, which generates fully interpretable and reproducible vectors representing a wide range of linguistic features in text \cite{okulska2023stylometrix}. Prior work shows that StyloMetrix performs well on persuasion detection with classical machine learning \cite{modzelewski-etal-2023-dshacker, modzelewski2024bilingual}, underscoring its suitability for studying the linguistic features of persuasive texts. In our analysis, we examined the full range of linguistic features offered by open-source StyloMetrix.


Our main contributions are summarized as:
\begin{itemize}[nosep, leftmargin=*]
    \item We introduce Persuaficial, a novel persuasion benchmark with approximately 65k multilingual texts generated via four controllable approaches with four LLMs.
    \item We are the first to investigate whether persuasive text generated by LLMs is harder to detect than human-written persuasive text. We conduct this analysis across four LLMs and 16 controllable generation settings, providing a comprehensive evaluation of detection difficulty across diverse persuasive text generation approaches.
    \item Our work is the first to characterize the linguistic differences between LLM-generated and human-written English persuasive texts. Our analysis encompasses 196 distinct linguistic features.
\end{itemize}
We release our codebase, dataset and all prompts\footnote{\href{https://github.com/ArkadiusDS/Persuaficial}{https://github.com/ArkadiusDS/Persuaficial}}.


\section{Human Persuasion Datasets}
\label{sec:human_datasets}
In our analysis, we employed three well-established datasets that had been previously annotated by humans. Using multiple datasets mitigates potential bias that could arise from relying on a single source and ensures a broader coverage of persuasion phenomena. We selected datasets that are widely used and cited in persuasion research. Below, we describe the human-created datasets used in our study:
\begin{itemize}[nosep, leftmargin=*]
    \item \textbf{SemEval 2023 Task 3 Dataset}: A multilingual, multifaceted collection of online news articles annotated with various persuasion techniques on paragraph level \cite{piskorski2023multilingual}. Its taxonomy and dataset are widely adopted within the NLP community for persuasion research \cite{barron2024overview, dimitrov2024semeval, modzelewski2025pcot}. This dataset was introduced as part of SemEval 2023 Task 3 on persuasion detection \cite{piskorski-etal-2023-semeval}. The extended taxonomy was also applied to annotate a parliamentary debates and social media dataset presented at the SlavicNLP workshop \cite{piskorski-etal-2025-slavicnlp}.
    \item \textbf{DIPROMATS 2024 Task 1 Dataset}: A dataset consisting of posts from the X platform (former Twitter) used for the DIPROMATS 2024 shared task including propaganda detection \cite{moral2024overview, moral2023overview}. The dataset contains messages from diplomats and authorities of major world powers, including China, the United States, Russia, and the European Union. DIPROMATS 2024 was part of IberLEF, an annual Spanish shared evaluation campaign \cite{chiruzzo2024overview}.
    \item \textbf{ChangeMyView}: A dataset derived from the Reddit \emph{ChangeMyView} discussion community. Dataset contains 3,051 conversations in which the persuader tries to convince the persuadee to change their mind \cite{tan2016winning}. It is commonly used in persuasion research \cite{ziems2024can, wei-etal-2016-post, dutta2020changing, labruna-etal-2026-detecting}.
\end{itemize}
Together, these datasets provide diverse perspectives on human persuasion, covering different sources and communicative contexts. Consequently, our analysis captures diverse persuasion understanding, which improves the generalizability of our findings. For further justification of our dataset choices, please refer to Appendix \ref{app:human_dataset_preparation}.

\section{Persuaficial: Artificially Generated Persuasion Dataset} 
\label{sec:dataset_construction}

In this section, we introduce a new dataset, \textbf{Persuaficial}, which contains persuasion texts generated artificially using large language models.

\subsection{Persuasive Text Generation Approaches}
Inspired by \citet{chencan}, which explores methods for generating synthetic misinformation, we developed a controllable persuasion generation \(G_P\), in which \(P\) denotes one of four approaches for generating persuasive text: \textit{Paraphrasing}, \textit{Rewriting with Subtle Persuasion}, \textit{Rewriting with Intensified Persuasion}, and \textit{Open-ended} generation.


Each persuasive text was generated under controlled prompting conditions. Specifically, prompts included human-annotated persuasive texts together with instructions to paraphrase or rewrite the input while preserving, strengthening, or softening its persuasive effect. For the open-ended generation, following the approach of \citet{chencan}, the model was provided with concise summaries of the corresponding human-annotated persuasive examples and prompted to generate persuasive text freely based on summary. This procedure was applied to all selected instances across the chosen texts, using identical prompt templates for all languages. For non-English cases, we appended instructions specifying the target language of generation.

Controlling the generation process through explicit instructions is crucial for our study, as it ensures that the resulting LLM-generated persuasive texts remain semantically comparable to human-written texts. This comparability is essential for reliable evaluation of both persuasion detection performance and linguistic differences between human- and AI-generated persuasive texts.

All approaches for generating synthetic text with persuasion, along with example prompts, are included in Table \ref{tab:persuasion_generation_approaches} (details about our prompts presented in the Appendix \ref{app:prompt_templates_data_generation}).

\begin{table}[t]
\centering
\footnotesize
\renewcommand{\arraystretch}{1.05} 
\begin{tabular}
{p{1.45cm}<{\raggedright\arraybackslash} 
p{5cm}<{\raggedright\arraybackslash} 
}
\hline
\textbf{Approach} & \textbf{Prompt Example} \\ 

\hline
Paraphrased Persuasion & 
\redhl{Given a passage, please paraphrase it.} The original content meaning and level of persuasion must be strictly preserved. The passage is: \bluehl{<passage>}

\\
\hline
Rewritten Persuasion (Subtle Effect) & 
Given a passage, \redhl{please rewrite it to make its persuasiveness more subtle and harder to detect}. The original content and meaning should be the same. The passage is: \bluehl{<passage>}
\\
\hline
Rewritten Persuasion (Intensified Effect) & 
Given a passage, \redhl{please rewrite it so that the persuasion present in the content is refined for a stronger persuasive effect}. The original content and meaning should be the same. The passage is: \bluehl{<passage>}
\\
\hline
Open-Ended Persuasion & 

\redhl{Given a sentence, please write a short piece of text. The short text must contain persuasion.} The sentence is: \brownehl{<sentence>}
\\
\hline
\end{tabular}
\caption{Overview of four approaches used for generating persuasive texts with LLMs. Each method represents a distinct level of control over persuasive strength and content nature.}
\label{tab:persuasion_generation_approaches}
\end{table}

\subsection{Persuaficial Dataset Construction}

Persuaficial is an AI-generated persuasion dataset constructed using multiple LLMs and diverse generation approaches. For \textit{Paraphrasing}, \textit{Rewriting (Subtle Effect)}, and \textit{Rewriting (Intensified Effect)}, we sample 1,000 texts from three real-world persuasion datasets (described in Section~\ref{sec:human_datasets}). Each sample includes 500 texts annotated as persuasive and 500 labeled as non-persuasive\footnote{The only exception was German: the corpus contained only 420 non-persuasive texts, so we included all of them and randomly sampled 580 persuasive texts.}.
Each selected persuasive text is treated as \bluehl{<passage>} (see Table~\ref{tab:persuasion_generation_approaches}) and serves as input for the generation method. For \textit{Open-ended} generation, we first summarize each selected persuasive text into a factual statements. We use the resulting \brownehl{<sentence>} (see Table~\ref{tab:persuasion_generation_approaches}) for the generation of persuasive synthetic text.

We employ open-weight and proprietary LLMs for dataset construction. The open models include \textit{Gemma~3~27b~it} and \textit{Llama~3.3~70B}. The commercial models are \textit{Gemini~2.0~Flash} and \textit{GPT~4.1~Mini}. Additional details on dataset creation and hyperparameter settings are provided in Appendix~\ref{app:dataset_generation}. Details about the models, APIs used, and the rationale for model selection is available in Appendix~\ref{app:llm_used_selection_rationale}.

\subsection{Persuaficial Quality Evaluation}

\paragraph{Pre-Generation Quality Evaluation.}

As mentioned, for the \textit{Open-ended} generation setting, we first summarize each selected persuasive text into a short \brownehl{<sentence>}. To ensure these LLM-generated \brownehl{<sentence>}s accurately reflect the content of the corresponding human texts, we conducted a human evaluation following explicit and rigorous annotation guidelines (see Appendix~\ref{app:pre_annotation_guidelines}). 

Two annotators were first trained by one of the authors, who has prior experience in annotation. A small training sample of 50 English sentences was selected, and the annotators independently applied the annotation guidelines, with opportunities to discuss their decisions. After completing the training phase, they reviewed and discussed their independent evaluations to align their understanding of the guidelines. The annotations from this training phase were excluded from further evaluation.

For the final evaluation, a sample of 200 English \brownehl{<sentence>}s was selected. Two independent annotators assessed each \brownehl{<sentence>} for factual correspondence. We then computed the accuracy of the LLM-generated \brownehl{<sentence>}s, considering as positive only those instances where both annotators independently agreed that a sentence was factual. The resulting accuracy of the LLM generation process was about 91.2\%, suggesting that LLMs may be effective at transforming texts into a short factual statements.
Moreover, most mismatches between the generated \brownehl{<sentence>}s and source texts were minor in nature, e.g., converting an exclamatory formulation (“\textit{Introduce the law!}”) into a declarative one (“\textit{The law will be introduced.}”). 
This result suggests that the generated factual sentences are of generally high quality and are unlikely to negatively impact the overall quality of the resulting Persuaficial dataset and our experiments. 


\paragraph{Post-Generation Quality Evaluation.}

To ensure that the final Persuaficial dataset meets the intended goal of containing persuasive content produced by LLMs under controlled prompting conditions, we conducted a multi-stage post-generation quality evaluation. While the pre-generation evaluation ensured that the factual sentences for \textit{Open-ended} approach were valid, the post-generation evaluation verifies whether the LLM-generated persuasive variants are (1) faithful to the target factual content, (2) persuasive, and (3) faithful to the instruction from persuasion generation approach.

We adopted a two-layer rigorous verification design that separates basic validity checking from persuasion-specific judging. 
We verified 400 generated English texts, each independently annotated by two trained annotators following detailed instructions (Appendix~\ref{app:post_annotation_guidelines}). 
As a result, we report an overall accuracy metric, defined as the proportion of generated texts unanimously annotated as valid by two annotators, where validity required that all three criteria received a positive annotation.

Due to the conservative requirement that all three criteria be jointly satisfied, the overall accuracy is 88.2\%. Most invalid cases involve only minor factual deviations rather than substantive inconsistencies. When considering persuasion-related criteria alone, accuracy grows to 97.69\%, indicating that LLMs reliably generate persuasive text and justifying the use of this data for subsequent comparisons between human- and AI-generated persuasive texts.



\subsection{Data Statistics}

For each language, we sampled 1,000 human-written passages from the original datasets, including 500 persuasive texts. For each persuasive example, we generated AI-counterparts using four LLMs and four generation approaches (4 models × 4 approaches = 16 generation configurations). This resulted in approx. 24,000 texts in English and 41,000 texts for the non-English languages. Overall, Persuaficial is a multilingual corpus of about 65,000 texts. Table \ref{tab:persuaficial_data_stats} presents basic statistics for our dataset. Detailed statistics in Appendix \ref{appendix:add_persuaficial_stats}.

\begin{table*}[!ht]
\centering
\scriptsize
\begin{tabular}{lccccc}
\hline
\textbf{Data Statistic} & 
\textbf{Human-written} & \textbf{Paraphrasing Generation} & 
\multicolumn{2}{c}{\textbf{Rewriting Generation}} & 
\textbf{Open-ended Generation} \\ 
 & & & \textbf{Subtle Persuasion} & \textbf{Intensive Persuasion} & \\ \hline
Average No. of Words & 72 & 79 & 81 & 87 & 65 \\
Average No. of Characters & 452 & 538 & 568 & 605 & 450 \\
\hline
\end{tabular}
\caption{Basic statistics for human-written and for Persuaficial dataset, including LLM-generated persuasive texts.}
\label{tab:persuaficial_data_stats}
\end{table*}

\section{Automatic Detection of Human and AI-Generated Persuasion}
\label{sec:persuasion_classification}

\subsection{Experimental Setup}

For our experiments, we use Persuaficial, which comprises artificially generated persuasive texts. Moreover, we use human-written counterparts. Each experiment uses data that is balanced across persuasive and non-persuasive classes.

For automatic persuasion detection, we employed four LLMs: \textit{GPT-4.1 Mini}, \textit{Gemini 2.0 Flash}, \textit{Gemma 3 27B Instruct}, and \textit{Llama 3.3 70B}. To ensure as deterministic outputs as possible, we set the temperature to 0 during classification. Since our goal is to detect persuasion, we formulate the task as a binary classification problem.  All classifications were performed in a zero-shot setting. This approach aligns with our research question (RQ1). Moreover, studies show that zero-shot detection with modern LLMs (e.g., GPT-4) can outperform supervised models such as BERT on binary classification tasks \cite{pelrine2023towards, bang2023multitask, hassan2020political}. Furthermore, \citet{lucas2023fighting} and \citet{modzelewski2025pcot} report that while fine-tuning BERT on multiple datasets results in poor generalization to unseen data, zero-shot LLMs maintain strong cross-domain performance. We evaluate persuasion detection performance using the F\textsubscript{1} score. Further details supporting reproducibility, including the LLM classifiers setup and the prompt templates used for persuasion detection, are provided in Appendix~\ref{appendix:cls_details}.

\subsection{Results on English Datasets}

Table \ref{tab:result_avg_english} reports F\textsubscript{1} scores for persuasion detection on three human-written balanced samples and their LLM-generated counterparts produced using four generation approaches.

On the \textit{Paraphrasing} subset of our Persuaficial dataset, F\textsubscript{1} scores are only marginally lower than those for human-written texts (on average 0.67\% lower), indicating that paraphrasing preserves a similar level of difficulty for persuasion detection across human and generated texts. In contrast, \textit{Rewriting (Intensified)} and \textit{Open-ended} subsets yield the highest F\textsubscript{1} scores. On average, persuasion is 9.75\% easier to detect in open-ended scenario and 5.33\% easier when persuasion is intensified. This makes open-ended generated persuasive texts the easiest setting for LLM-based detection. We hypothesize that models tend to over-express explicit persuasive cues when prompted to generate persuasive text freely or while intensifying persuasion, which in turn makes these texts more easily detectable. The opposite pattern emerges for \textit{Rewriting (Subtle persuasion)}, where F\textsubscript{1} scores drop substantially, by 20.42\% on average. This suggests that reducing overt persuasive markers makes persuasion significantly harder to detect, even for strong LLM detectors. Importantly, these patterns are highly consistent across datasets and across all detector models. This may indicate that the effects generalize across domains and are independent of the specific LLM used for detection.

In summary, addressing RQ1, the detectability of LLM-generated persuasive text depends on the generation approach: texts produced via open-ended and intensified persuasion are easier to detect, whereas subtly persuasive generations remain substantially more challenging for current LLM-based detectors.
Detailed results in Appendix \ref{app:human_machine_persuasion_text_class_results_english}.

\begin{table*}[!ht]
\centering
\scriptsize
\begin{tabular}{lccccc}
\hline
\textbf{Classifier Models} & 
\textbf{Human-written} & \textbf{Paraphrasing Generation} & 
\multicolumn{2}{c}{\textbf{Rewriting Generation}} & 
\textbf{Open-ended Generation} \\ 
 & & & \textbf{Subtle Persuasion} & \textbf{Intensive Persuasion} & \\ \hline
 \multicolumn{6}{l}{\textit{\textbf{Sample of Persuaficial generated based on: SemEval 2023 data}}} \\
GPT 4.1 Mini & 0.7398 & \perc{0.7398}{0.7007} & \perc{0.7398}{0.4031} & \perc{0.7398}{0.8148} & \perc{0.7398}{0.8964} \\
Llama 3.3 70B & 0.7459 & \perc{0.7459}{0.7207} & \perc{0.7459}{0.4577} & \perc{0.7459}{0.8111} & \perc{0.7459}{0.8741} \\
Gemma 3 27b it & 0.7572 & \perc{0.7572}{0.7592} & \perc{0.7572}{0.6453} & \perc{0.7572}{0.8208} & \perc{0.7572}{0.8562} \\
Gemini 2.0 Flash & 0.7551 & \perc{0.7551}{0.7540} & \perc{0.7551}{0.6522} & \perc{0.7551}{0.7950} & \perc{0.7551}{0.8117} \\
\hline
\multicolumn{6}{l}{\textit{\textbf{Sample of Persuaficial generated based on: DIPROMATS 2024 data}}} \\
GPT 4.1 Mini & 0.7567 & \perc{0.7567}{0.7461} & \perc{0.7567}{0.4962} & \perc{0.7567}{0.8100} & \perc{0.7567}{0.8666} \\
Llama 3.3 70B & 0.7471 & \perc{0.7471}{0.7362} & \perc{0.7471}{0.5696} & \perc{0.7471}{0.7860} & \perc{0.7471}{0.8292} \\
Gemma 3 27b it & 0.7473 & \perc{0.7473}{0.7460} & \perc{0.7473}{0.6349} & \perc{0.7473}{0.7782} & \perc{0.7473}{0.7994} \\
Gemini 2.0 Flash & 0.7518 & \perc{0.7518}{0.7427} & \perc{0.7518}{0.6664} & \perc{0.7518}{0.7640} & \perc{0.7518}{0.7680} \\
\hline
\multicolumn{6}{l}{\textit{\textbf{Sample of Persuaficial generated based on: ChangeMyView data}}} \\
GPT 4.1 Mini & 0.6233 & \perc{0.6233}{0.6356} & \perc{0.6233}{0.4906} & \perc{0.6233}{0.6739} & \perc{0.6233}{0.7148} \\
Llama 3.3 70B & 0.6517 & \perc{0.6517}{0.6488} & \perc{0.6517}{0.5536} & \perc{0.6517}{0.6691} & \perc{0.6517}{0.6831} \\
Gemma 3 27b it & 0.6644 & \perc{0.6644}{0.6708} & \perc{0.6644}{0.6334} & \perc{0.6644}{0.6809} & \perc{0.6644}{0.6843} \\
Gemini 2.0 Flash & 0.6671 & \perc{0.6671}{0.6662} & \perc{0.6671}{0.6294} & \perc{0.6671}{0.6740} & \perc{0.6671}{0.6770} \\
\hline
\end{tabular}
\caption{F\textsubscript{1} scores for persuasion detection on English data. The first column reports performance on human-annotated texts. The remaining columns show performance on LLM-generated texts. For generated data, each value represents the average F\textsubscript{1} score obtained from classification of texts generated by four different LLMs. Detailed results without averaging F\textsubscript{1} scores in Appendix \ref{app:human_machine_persuasion_text_class_results_english}.}
\label{tab:result_avg_english}
\end{table*}

\subsection{Results on Non-English Datasets}

Table \ref{tab:result_avg_non_english} shows persuasion detection results for German, French, Italian, Polish, and Russian. The patterns observed in English hold consistently across all languages and classifiers. Paraphrasing preserves a difficulty level similar to human-written texts, whereas intensified rewriting and open-ended generation yield the highest F\textsubscript{1} scores. Open-ended generation frequently yields F\textsubscript{1} scores above 0.9, indicating that persuasive text is easiest to detect in this setting.
In contrast, subtle rewriting causes the largest drop in performance. These trends suggest that generation approaches influence persuasion detectability, with effects that may generalize across languages. Detailed results in Appendix \ref{app:human_machine_persuasion_text_class_results_non_english}.


\begin{table*}[!ht]
\centering
\scriptsize
\begin{tabular}{lccccc}
\hline
\textbf{Classifier Models} & 
\textbf{Human-written} & \textbf{Paraphrasing Generation} & 
\multicolumn{2}{c}{\textbf{Rewriting Generation}} & 
\textbf{Open-ended Generation} \\ 
 & & & \textbf{Subtle Persuasion} & \textbf{Intensive Persuasion} & \\ \hline
 \multicolumn{6}{l}{\textit{  \textbf{German}}}\\
GPT 4.1 Mini & 0.7203 & \perc{0.7203}{0.7207} & \perc{0.7203}{0.4410} & \perc{0.7203}{0.8456} & \perc{0.7203}{0.9414} \\ 
Llama 3.3 70B & 0.7361 & \perc{0.7361}{0.7248} & \perc{0.7361}{0.4398} & \perc{0.7361}{0.8474} & \perc{0.7361}{0.9345} \\ 
Gemma 3 27b it & 0.7655 & \perc{0.7655}{0.7763} & \perc{0.7655}{0.6664} & \perc{0.7655}{0.8512} & \perc{0.7655}{0.9004} \\ 
Gemini 2.0 Flash & 0.7903 & \perc{0.7903}{0.7880} & \perc{0.7903}{0.6905} & \perc{0.7903}{0.8385} & \perc{0.7903}{0.8591} \\ 
\hline
 \multicolumn{6}{l}{\textit{  \textbf{French}}}\\
GPT 4.1 Mini & 0.7505 & \perc{0.7505}{0.7454} & \perc{0.7505}{0.4290} & \perc{0.7505}{0.8456} & \perc{0.7505}{0.9251} \\ 
Llama 3.3 70B & 0.7605 & \perc{0.7605}{0.7450} & \perc{0.7605}{0.4527} & \perc{0.7605}{0.8432} & \perc{0.7605}{0.9172} \\ 
Gemma 3 27b it & 0.7827 & \perc{0.7827}{0.7866} & \perc{0.7827}{0.6587} & \perc{0.7827}{0.8476} & \perc{0.7827}{0.8824} \\ 
Gemini 2.0 Flash & 0.7812 & \perc{0.7812}{0.7860} & \perc{0.7812}{0.6800} & \perc{0.7812}{0.8314} & \perc{0.7812}{0.8418} \\ 
\hline
 \multicolumn{6}{l}{\textit{  \textbf{Italian}}}\\
GPT 4.1 Mini & 0.7471 & \perc{0.7471}{0.7330} & \perc{0.7471}{0.4246} & \perc{0.7471}{0.8428} & \perc{0.7471}{0.9195} \\ 
Llama 3.3 70B & 0.7584 & \perc{0.7584}{0.7172} & \perc{0.7584}{0.4285} & \perc{0.7584}{0.8420} & \perc{0.7584}{0.9161} \\ 
Gemma 3 27b it & 0.7659 & \perc{0.7659}{0.7804} & \perc{0.7659}{0.6610} & \perc{0.7659}{0.8399} & \perc{0.7659}{0.8686} \\ 
Gemini 2.0 Flash & 0.7986 & \perc{0.7986}{0.7938} & \perc{0.7986}{0.6781} & \perc{0.7986}{0.8301} & \perc{0.7986}{0.8408} \\ 
\hline
 \multicolumn{6}{l}{\textit{  \textbf{Polish}}}\\
GPT 4.1 Mini & 0.7330 & \perc{0.7330}{0.7060} & \perc{0.7330}{0.4580} & \perc{0.7330}{0.8483} & \perc{0.7330}{0.9367} \\ 
Llama 3.3 70B & 0.7676 & \perc{0.7676}{0.7389} & \perc{0.7676}{0.5041} & \perc{0.7676}{0.8518} & \perc{0.7676}{0.9206} \\ 
Gemma 3 27b it & 0.7728 & \perc{0.7728}{0.7783} & \perc{0.7728}{0.6919} & \perc{0.7728}{0.8427} & \perc{0.7728}{0.8834} \\ 
Gemini 2.0 Flash & 0.7733 & \perc{0.7733}{0.7732} & \perc{0.7733}{0.7018} & \perc{0.7733}{0.8101} & \perc{0.7733}{0.8217} \\ 
\hline
 \multicolumn{6}{l}{\textit{  \textbf{Russian}}}\\
GPT 4.1 Mini & 0.7246 & \perc{0.7246}{0.7073} & \perc{0.7246}{0.4392} & \perc{0.7246}{0.8242} & \perc{0.7246}{0.9017} \\ 
Llama 3.3 70B & 0.7408 & \perc{0.7408}{0.7164} & \perc{0.7408}{0.4312} & \perc{0.7408}{0.8324} & \perc{0.7408}{0.9086} \\ 
Gemma 3 27b it & 0.7360 & \perc{0.7360}{0.7416} & \perc{0.7360}{0.6128} & \perc{0.7360}{0.8098} & \perc{0.7360}{0.8562} \\ 
Gemini 2.0 Flash & 0.7683 & \perc{0.7683}{0.7616} & \perc{0.7683}{0.6795} & \perc{0.7683}{0.7877} & \perc{0.7683}{0.8019} \\ 
\hline
\end{tabular}
\caption{F\textsubscript{1} scores for persuasion detection on non-English data samples. The first column reports performance on human-annotated texts. The remaining columns show performance on LLM-generated texts. For generated data, each value represents the average F\textsubscript{1} score obtained from classification of texts generated by four different LLMs. Detailed results without averaging F\textsubscript{1} scores in Appendix \ref{app:human_machine_persuasion_text_class_results_non_english}}
\label{tab:result_avg_non_english}
\end{table*}

\section{Linguistic Differences Between Machine and Human Persuasion}

\label{sec:persuasion_linguistic_differences}
In this section, we investigate the linguistic differences between human-written and AI-generated persuasive texts. We focus on English due to the limited availability of high-quality datasets annotated with persuasion in other languages. While the SemEval 2023 Task 3 data provides a multilingual resource \cite{srba2024survey, piskorski-etal-2023-multilingual}, relying solely on it could introduce dataset-specific biases. To mitigate this, we use English part of Persuaficial and human-written counterparts from three well-established datasets. 

In our analysis, we focus on linguistic features that encompass morphosyntactic dimensions (e.g., inflection, POS distribution), syntax, lexis, and punctuation, among others.




\subsection{Our Approach for Linguistic Analysis}
To investigate the linguistic differences between human-written persuasive texts and LLM-generated persuasive texts, we adopt an explainable, feature-based analysis grounded in stylometry.
Our objective is to identify the linguistic features that most strongly differentiate LLM-generated persuasive texts from human-written ones.
For each linguistic feature, we compare the distributions of the two groups using effect-size-based analysis together with significance testing. Effect sizes quantify the magnitude of the difference between human and AI-generated texts for each feature \cite{frey2021sage}, while significance testing evaluates whether these distributional differences are statistically meaningful. We restrict our analysis to persuasive texts and aim to characterize differences between human and AI-generated persuasive text, without attempting to isolate persuasion-specific features from broader AI-human writing differences.

\subsection{Experimental Setup}
We first represent each persuasive human text and its AI-generated counterpart using StyloMetrix \cite{okulska2023stylometrix}. For each text, we calculate a 196-dimensional vector of linguistic features. This results in a tabular representation, where each row corresponds to a text encoded by its computed linguistic features. We utilize StyloMetrix, because it is an open-source tool that provides fully interpretable, linguistically grounded feature representations. Moreover, prior work has demonstrated its effectiveness for persuasion detection using classical machine learning models \cite{modzelewski-etal-2023-dshacker, modzelewski2024bilingual}, confirming its suitability for analyzing the linguistic characteristics of persuasive texts. Finally, to further justify our choice, we show that StyloMetrix features with classical machine learning can distinguish human-written from AI-generated persuasive texts in Appendix~\ref{app:stylometrix_justification}.

For each linguistic feature $j$, we directly compare its distribution in human-written and LLM-generated persuasive texts, conducting this analysis separately for each generation approach. We utilize paired Cohen’s $d$ statistic
which is a type of effect size measure used to represent the magnitude of differences between two groups on a given variable \cite{frey2021sage}. We report the paired version of Cohen’s $d$, as our analysis is based on matched pairs consisting of AI-generated persuasive texts and their corresponding human-written counterparts.


In our calculation, we firstly compute the element-wise difference between the generated and real feature values across all $n=1,500$ persuasive text pairs (500 per dataset, across three datasets):

\begin{equation}
\Delta_j^{i} = g_j^{i} - r_j^{i}, \quad i = 1, \ldots, 1500
\end{equation}

where $g_j^{i}$ denotes the value of feature $j$ for the $i$-th AI-generated text and $r_j^{i}$ denotes that of the corresponding human-written text. To measure the magnitude of the shift introduced by the language model, we compute paired Cohen's $d$:

\begin{equation}
d_j = \frac{\bar{\Delta}_j}{s_{\Delta_j}}
\label{eq:cohensd}
\end{equation}
where
\begin{equation}
\bar{\Delta}_j = \frac{1}{n}\sum_{i=1}^{n} \Delta_j^{i}
\label{eq:mean_diff}
\end{equation}
 is the mean of the paired differences for feature $j$, and
 \begin{equation}
s_{\Delta_j} = \sqrt{\frac{1}{n-1}\sum_{i=1}^{n}(\Delta_j^{i} - \bar{\Delta}_j)^2}
\label{eq:std_diff}
\end{equation}
  is the standard deviation of the paired differences. This procedure is repeated independently for each feature $j$, each generating model, and each generation technique.

A positive Cohen's $d$ indicates the AI-generated text exhibits higher values of a given feature compared to the human original, while a negative value indicates a decrease. 

To evaluate whether feature distributions significantly differ between human and AI-generated texts, we perform Wilcoxon signed-rank tests \cite{wilcoxon1945individual}, a non-parametric paired test appropriate for comparing matched text pairs \cite{peyrard2021better, dror2018hitchhiker}. Wilcoxon signed-rank tests have seen widespread adoption in the NLP community \cite{karwa2025disentangling, zhou-etal-2025-socialeval, ciaccio2025evaluating}, including in studies comparing feature distributions \cite{stodden2020multi}. See Appendix \ref{app:wilcoxon_signed-rank} for more details regarding our statistical analysis.

\begin{figure}[ht!]
  \fbox{
    \begin{minipage}{0.94\columnwidth}
      \scriptsize 
    \purehl{G\_ACTIVE} The proportion of verbs in the text used in the active voice.

    \purehl{L\_ADV\_SUPERLATIVE} Measures how often superlative adverbial (and some adjective-as-adverb) forms appear in a text.

    \purehl{L\_ADV\_COMPARATIVE} The proportion of tokens that are adverbs used in comparative degree (e.g., "more", "less", or marked comparative forms)

    \purehl{L\_FUNC\_A} The proportion of tokens in a text that are function words.
      
    \purehl{L\_CONT\_T} The proportion of unique content-word forms in relative to total tokens.
      
    \purehl{L\_CONT\_A} The proportion of tokens in a text that are content words.

    \purehl{L\_PUNCT\_COM} Comma incidence measures frequency of commas relative to text length.
      
    \purehl{L\_PUNCT\_DASH} Measures the density of dashes within a text.

    \purehl{L\_PUNCT\_DOT} Measures the incidence of periods (dots) relative to the total number of words.

    
    \purehl{L\_PLURAL\_NOUNS} Measures the density of plural nouns within a text.

     \purehl{LTOKEN\_RATIO\_LEM} (ST\_TYPE\_TOKEN\_RATIO\_LEMMAS) The ratio of unique lemmas to total tokens.

    
    \purehl{POS\_ADJ} The proportion of tokens in the text that are adjectives, indicating the level of descriptiveness.

    \purehl{POS\_NOUN} The proportion of tokens in the text that are nouns.
 
    \purehl{POS\_PRO} The proportion of tokens in the text that are pronouns.

    \purehl{PS\_CAUSE} Measures the incidence of linking words and phrases related to cause and purpose.

     \purehl{SENT\_D\_NP} (ST\_SENT\_D\_NP) Measures the average proportion of noun phrase (NPs) tokens relative to sentence length, averaged over all sentences in a document.

   \purehl{SENT\_D\_PP} (ST\_SENT\_D\_PP) Measures the average proportion of tokens that belong to prepositional phrases (PPs) in each sentence, averaged over all sentences in the document.
     
     \purehl{SENT\_D\_VP} (ST\_SENT\_D\_VP) Measures the average proportion of tokens in a sentence that are not marked with a verb tense, relative to total sentence length, averaged over all sentences in the document.
     
    \purehl{SENT\_ST\_DIFFER} (ST\_SENT\_DIFFERENCE) Quantifies syntactic variation between consecutive sentences by comparing their dependency label sets and averaging over the document.
    
    \purehl{SENT\_ST\_WPERSENT} Indicates the normalized difference between the total number of tokens and the number of sentences in a document (a proxy for sentence length).

     \purehl{ST\_REPET\_WORDS} (ST\_REPETITIONS\_WORDS) Measures the level of lexical repetition by computing the proportion of repeated word tokens in a text normalized by total token count.

     \purehl{SY\_EXCLAMATION} Measures the proportion of unique word tokens that appear in exclamatory sentences relative to all tokens in the text.
     
    \purehl{SY\_IMPERATIVE} Measures the proportion of unique alphabetic words that appear in sentences classified as imperative, relative to all tokens in the document.

     \purehl{SY\_INV\_PATTERNS} (SY\_INVERSE\_PATTERNS) A syntactic feature that measures the frequency of inverted sentence structures within a text.

      \purehl{SY\_NARRATIVE} Measures the proportion of tokens in declarative sentences relative to all tokens.

    \purehl{VF\_INFINITIVE} A syntactic feature that measures the proportion of infinitive verb forms.

      \purehl{VT\_MIGHT} Measures the frequency of "might" in a text.

    \end{minipage}
  }
    \caption{StyloMetrix features that exhibit some of the strongest discriminative power for distinguishing human-written persuasive text from persuasive content generated by LLMs.}
  \label{fig:ling_features_definitions}
\end{figure}

\subsection{Results and Analysis}
We computed Cohen’s d values for 196 linguistic features across four generation approaches and  four LLMs utilized to generate texts for our Persuaficial dataset. Table \ref{tab:stylometric_features} sorts top linguistic features by the absolute Cohen's d ($|d_j|$) values per model and generation approach ($G_P$). Our analysis and discussion is based on the twenty features with the largest $|d_j|$ for each model and generation strategy (more detailed tables available in Appendix \ref{app:ling_diff_for_add_results}). Statistical analysis using Wilcoxon signed-rank tests confirmed that all twenty features for each scenario exhibit significant distributional differences between human-written and AI-generated persuasive texts. Figure \ref{fig:ling_features_definitions} provides definitions of the key differentiating features.

High values of features such as L\_CONT\_T (the proportion of unique content-word forms relative to total tokens), LTOKEN\_RATIO\_LEM (the ratio of unique lemmas to total tokens), and L\_CONT\_A (the proportion of tokens that are content words) indicate that AI-generated texts tend to contain more varied content words and higher informational density per sentence. These patterns suggest that lexical diversity and content richness are characteristic markers of AI authorship. Similarly, low values for ST\_REPET\_WORDS are indicative of AI-generated persuasive texts, suggesting that reduced word repetition serves as a strong signal of LLM-generated text. A higher proportion of function words (L\_FUNC\_A) indicates that a persuasive text is likely human-written. This means that frequent use of grammatical connectors (such as articles, prepositions, pronouns, and auxiliary verbs) is a signal of human text. 
Furthermore, AI persuasive texts generated by \textit{Llama} tend to have a lower density of periods. However, certain punctuation marks, including commas (L\_PUNCT\_COM) and dashes (L\_PUNCT\_DASH), occur more frequently in AI texts.
The rarity of syntactically marked constructions, such as inversions (SY\_INV\_PATTERNS), is a distinguishing feature of AI text, as these complex syntactic patterns are more typical of human-written persuasive texts.

In AI-generated texts that intensify persuasion, comparative and superlative adverbs (L\_ADV\_COMPARATIVE with words like "more", "faster", and L\_ADV\_SUPERLATIVE with words like   "best", "worst") appear more frequently than in human-written texts. This suggests that AI strengthens persuasive language through the increased use of adverbial modifiers, highlighting a distinctive stylistic strategy in LLM-generated intensified texts.

In AI-generated texts that aim to make persuasion more subtle, modal verbs such as "might" (VT\_MIGHT) occur more frequently than in human-written texts. Similarly, narrative framing (SY\_NARRATIVE), defined as the proportion of tokens in declarative sentences relative to all tokens, is more prevalent in AI subtle rewritings. These patterns may indicate that AI softens persuasion by using modal hedges and favors neutral declarative constructions over exclamatory sentences or rhetorical questions more often than humans do.

Open-ended AI-generated texts exhibit a consistent linguistic profile characterized by high lexical diversity and elevated content-word density (e.g., L\_CONT\_T, L\_CONT\_A, LTOKEN\_RATIO\_LEM). AI systems also show substantially lower function-word usage and reduced lexical repetition. In addition, AI-generated texts rely more heavily on imperative and infinitival constructions while avoiding marked syntactic patterns such as inversions, which may result in structurally simpler and more schematic syntax.

\begin{table*}
\centering
\scriptsize
\begin{tabular}{lc|lc|lc|lc}
\hline
\multicolumn{2}{l|}{\textit{Paraphrasing}} & \multicolumn{2}{l|}{\textit{Rewriting for Subtle Effects}} &
\multicolumn{2}{l|}{\textit{Rewriting for Intensified Effects}} & \multicolumn{2}{l}{\textit{Open-ended} } \\
[2pt]
Feature Name & $d_j$ & Feature Name & $d_j$ & Feature Name & $d_j$ & Feature Name & $d_j$ \\
\hline
\noalign{\vskip 1pt}
\multicolumn{8}{l}{\textit{Generating Model: GPT 4.1 Mini}} \\
\noalign{\vskip 1pt}
 LTOKEN\_RATIO\_LEM& 0.92 &  LTOKEN\_RATIO\_LEM & 0.94 & L\_CONT\_T & 1.18 & L\_CONT\_T & 1.41  \\
L\_CONT\_T & 0.82 &  L\_CONT\_T & 0.82 & LTOKEN\_RATIO\_LEM & 1.09 & L\_CONT\_A & 1.23   \\
 ST\_REPET\_WORDS & -0.75 & ST\_REPET\_WORDS & -0.76  & L\_CONT\_A & 1.02  & LTOKEN\_RATIO\_LEM & 1.08 \\
 L\_CONT\_A  & 0.65 &    L\_PLURAL\_NOUNS  &  0.66 & L\_FUNC\_A & -0.86 &  ST\_REPET\_WORDS & -0.85  \\
L\_FUNC\_A & -0.58  & L\_CONT\_A & 0.63 & ST\_REPET\_WORDS & -0.82 & SENT\_D\_NP & 0.76\\
\hline
\noalign{\vskip 1pt}
\multicolumn{8}{l}{\textit{Generating Model: Llama 3.3 70B}} \\
\noalign{\vskip 1pt}
SENT\_ST\_WPERSENT &  0.77  & SENT\_ST\_WPERSENT & 0.73 &  L\_CONT\_T & 0.95 &  VF\_INFINITIVE & 1.01  \\
SENT\_ST\_DIFFER & -0.77 & L\_CONT\_T & 0.70 & SENT\_ST\_WPERSENT & 0.88 &  SY\_IMPERATIVE & 0.89 \\
L\_CONT\_T & 0.76 & SENT\_D\_PP & 0.67 & L\_CONT\_A & 0.76 &  G\_ACTIVE & -0.70 \\
L\_PUNCT\_COM & 0.68  & SENT\_ST\_DIFFER & -0.64 & L\_ADJ\_POSITIVE & 0.75 & SENT\_D\_VP & 0.68\\
LTOKEN\_RATIO\_LEM  & 0.61 & L\_PROPER\_NAME & -0.61 &  SENT\_ST\_DIFFER & -0.73 & L\_OUR\_PRON & 0.60 \\
\hline
\noalign{\vskip 1pt}
\multicolumn{8}{l}{\textit{Generating Model: Gemma 3 27b it}} \\
\noalign{\vskip 1pt}
L\_CONT\_T & 0.87 &  L\_CONT\_T & 1.02 & L\_CONT\_T & 1.11 & SY\_IMPERATIVE & 1.25 \\
LTOKEN\_RATIO\_LEM & 0.73 & L\_CONT\_A & 0.87  & L\_FUNC\_A & -0.95 & L\_CONT\_T & 1.21 \\
L\_FUNC\_A & -0.70 &  LTOKEN\_RATIO\_LEM & 0.85 & L\_CONT\_A & 0.95 & L\_CONT\_A & 1.05 \\
L\_CONT\_A & 0.68  & L\_FUNC\_A & -0.76  & L\_ADJ\_POSITIVE & 0.83 & L\_FUNC\_A & -0.92 \\
ST\_REPET\_WORDS & -0.67 & ST\_REPET\_WORDS &  -0.67 & POS\_ADJ & 0.77 & PS\_CAUSE & -0.89 \\
\hline
\noalign{\vskip 1pt}
\multicolumn{8}{l}{\textit{Generating Model: Gemini 2.0 Flash}} \\
\noalign{\vskip 1pt}
L\_CONT\_T & 0.90 & L\_CONT\_T & 1.02 & L\_CONT\_T & 1.23 & L\_CONT\_T & 1.34 \\
LTOKEN\_RATIO\_LEM & 0.82 & L\_CONT\_A & 0.87 & L\_CONT\_A & 1.12 & L\_CONT\_A & 1.16 \\
L\_CONT\_A & 0.79 & LTOKEN\_RATIO\_LEM &  0.85 & L\_FUNC\_A & -0.98 & LTOKEN\_RATIO\_LEM & 0.97 \\
ST\_REPET\_WORDS & -0.72 & L\_FUNC\_A & -0.76 & LTOKEN\_RATIO\_LEM & 0.89 & SY\_IMPERATIVE & 0.93 \\
L\_FUNC\_A & -0.66 & ST\_REPET\_WORDS & -0.67 & L\_ADJ\_POSITIVE & 0.87  & ST\_REPET\_WORDS & -0.83 \\
\hline
\end{tabular}
\caption{Top linguistic features by absolute Cohen's d from four generation approaches for all generating LLMs.}
\label{tab:stylometric_features}
\end{table*}

\section{Related Work}
Research on the persuasive capabilities of generative AI spans multiple disciplines, including computer science as well as the social and complexity sciences \cite{duerr2021persuasive}. Recent progress in large language models has drawn attention to their potential for persuasion and related applications \cite{jin-etal-2024-persuading, rogiers2024persuasion}. Early studies by \citet{wang-etal-2019-persuasion} explored personalized persuasive dialogue systems designed to promote socially beneficial outcomes. Subsequent studies have investigated how individuals respond to persuasive machine-generated text and how they perceive its effectiveness \cite{karinshak2023working, bai2025llm, goldstein2024persuasive}. In addition, \citet{schoenegger2025large} explored whether LLMs can be more persuasive than humans, while \citet{pauli2025measuring} analyzed the extent to which LLMs are capable of generating persuasive language. 

Parallel work has investigated approaches for 
detecting persuasion. \citet{da2019fine} presented corpus of news annotated at the fragment level with eighteen persuasive techniques and proposed multi-granularity neural network to detect persuasion. \citet{piskorski2023multilingual} extended the taxonomy of techniques proposed by \citet{da2019fine} and presented a multilingual dataset. \citet{modzelewski-etal-2024-mipd} introduced a Polish disinformation dataset annotated with eleven manipulation techniques, defining manipulation as the intentional use of persuasion for malicious purposes. Moreover, persuasion detection was a task of different recognized workshops such SemEval or SlavicNLP \cite{da-san-martino-etal-2020-semeval, dimitrov-etal-2021-semeval, piskorski-etal-2023-semeval, piskorski2025slavicnlp}. 

To the best of our knowledge, no prior work has investigated whether LLM-generated persuasive texts are easier to automatically detect than human-written ones. Furthermore, existing research has not provided a linguistic analysis comparing LLM-generated and human persuasive content. 



\section{Conclusion}
In this work, we introduce \textbf{Persuaficial}, a multilingual benchmark of LLM-generated persuasive texts, comprising about 65,000 instances across English, German, French, Italian, Polish, and Russian. Our experiments show that the detectability of persuasion in generated texts strongly depends on the generation approaches: open-ended and rewriting with intensified persuasion increase detectable cues, whereas rewriting with subtle persuasion substantially reduces detection performance. These trends are consistent across languages and classifier models, indicating that generation prompts may shape the difficulty of persuasive text detection.

In addition, through a detailed analysis, we identify key linguistic differences between human and AI-generated persuasive texts. AI-generated texts tend to exhibit higher lexical diversity, increased content-word density, and lower function-word usage, while complex syntactic patterns are more characteristic of human persuasive writing. Text generation approaches further modulate these features: intensified persuasion amplifies adverbial modifiers, whereas subtle persuasion relies on modal hedges and declarative constructions.

Overall, our findings demonstrate that LLM-generated persuasive texts are not only linguistically distinct from human-written texts but also vary in detectability depending on the generation approach. \textbf{Persuaficial} provides a valuable resource for future research on automated persuasion detection, cross-lingual NLP, and the study of linguistic differences between human and AI-generated persuasive content.

\section*{Limitations}

While \textbf{Persuaficial} offers a large and diverse resource for studying AI persuasive text detection, several limitations remain. First, although the corpus covers six languages, our linguistic analysis focuses only on English. This decision stems from the limited availability of high-quality, human-written persuasive datasets in other languages and from the need for a controlled, comparable setup across human and AI-generated texts. Conducting the analysis exclusively on English avoids introducing dataset-specific biases that would arise from relying on a single non-English persuasion dataset, ensuring that the linguistic findings are not driven by characteristics of a particular corpus.

In our analysis, all classifier evaluations are conducted in a zero-shot setting, which aligns with the goals and research question of this work. We chose not to explore few-shot or alternatives, leaving these as directions for future research. Moreover, prior studies show that modern LLMs in zero-shot mode (e.g., GPT-4) can outperform supervised models such as BERT on binary classification tasks \cite{pelrine2023towards, bang2023multitask, hassan2020political}, and that fine-tuned BERT models may generalize poorly to out-of-domain data compared to zero-shot LLMs \cite{lucas2023fighting, modzelewski2025pcot}.

\section*{Ethics}

\paragraph{Dataset} The Persuaficial dataset consists of synthetic persuasive texts generated by large language models for research purposes. To construct the dataset, we relied exclusively on three established human-authored persuasion datasets that are either publicly available or were used with explicit permission from their original authors. No personally identifiable information is included in the generated persuasive content, and no attempt is made to identify or infer the authorship of individual texts.

All human-written and synthetic texts were used solely for academic research on persuasion detection and linguistic analysis. The generation process preserves the semantic content of the source material while producing novel text, thereby avoiding the reproduction of identifiable original passages. To promote transparency, reproducibility, and responsible reuse, the Persuaficial dataset will be released under the Creative Commons Attribution 4.0 (CC BY 4.0) license.

Crowdsourcing was not used at any stage of dataset creation or validation. All individuals involved in verifying the quality of Persuaficial were researchers or trained annotators with prior experience in persuasion or manipulation annotation. The verification process was conducted independently and remained free from political, institutional, or commercial influence.

\paragraph{Computational resources} The use of large language models is associated with non-trivial computational and environmental costs \cite{strubell2019energy}. In this work, we mitigate these costs by avoiding model training or fine-tuning and relying exclusively on inference with pre-existing models. All experiments were conducted through third-party APIs, and we did not directly manage or allocate the underlying computational infrastructure. As a result, the overall computational footprint of this study was limited to inference-time usage.

\section*{Acknowledgments}

Giovanni Da San Martino would like to thank the Qatar National Research Fund, part of Qatar Research Development and Innovation Council (QRDI), for funding this work  by grant NPRP14C0916-210015. 


\appendix

\section{Human Dataset - Rationale Behind Our Choice}
\label{app:human_dataset_preparation}

In our experiments, we adopt the concise definition of persuasion proposed by \citet{piskorski2023multilingual, piskorski2023news}: “\textit{Persuasive text is characterized by a specific use of language in order to influence the reader}”. We rely on this definition because it underpins the annotation guidelines of the SemEval 2023 Task 3 dataset, the largest publicly available resource for studying persuasion and one of the three human-written datasets used in our study. Consequently, we selected additional datasets that align well with this conceptualization of persuasion.

DIPROMATS 2024 \citep{moral2024overview}, which build on the SemEval task, offers data that is directly compatible with this definition and is therefore suitable for our experiments. Finally, \citet{piskorski2025slavicnlp} demonstrate that this definition can be effectively and reliably applied to debates and discussions, motivating our choice of the ChangeMyView dataset, comprising message exchanges between persuaders and persuadees, as an additional human-written source.

\section{Persuaficial Generation Process}
\label{app:dataset_generation}

\subsection{Human-written Text Sample Selection}
\label{app:sample_selection}

For each of the eight source datasets (DIPROMATS 2024, ChangeMyView, and SemEval 2023 Task 3 with six languages), we generated samples from a base of 500 persuasive human-written texts. By applying four generation approaches across four different LLMs, we produced 16 distinct machine-generated counterparts for every source text. Moreover, each sample contains 500 non-persuasive human-written texts.

\subsection{Experimental Setup for LLM Persuasion Generation Process}
\label{app:generating_persuasion_setup}
We employed four Large Language Models.
The open models include \textit{Gemma~3~27b~it} and \textit{Llama~3.3~70B}. The commercial models are \textit{Gemini~2.0~Flash} and \textit{GPT~4.1~Mini}.
To encourage more diverse, creative and less repetitive phrasing in the model outputs, we set the generation temperature to 0.8. Our choice of temperature for generating synthetic persuasive texts was directly informed by the settings used by \citet{chencan} in a related task involving misinformation generation.

\subsection{Prompt Templates for Persuaficial Dataset Creation}
\label{app:prompt_templates_data_generation}
Figures \ref{fig:prompt_paraphrased}, \ref{fig:prompt_subtle}, \ref{fig:prompt_intensified}, and \ref{fig:open_ended} show prompt templates used during the LLM persuasion generation process. In our prompts, we adopt the concise definition of persuasion proposed by \citet{piskorski2023multilingual, piskorski2023news}: “\textit{Persuasive text is characterized by a specific use of language in order to influence the reader}”.

\begin{figure*}[ht]
  \centering
  \begin{tcolorbox}[colback=ForestGreen!15!white, colframe=ForestGreen!50!black,
    width=\textwidth, boxrule=0.5pt, arc=3pt, auto outer arc,
    fonttitle=\bfseries, title=Generation approach: Paraphrasing Generation prompt]
\scriptsize
\textbf{System Prompt:} \\
You are an assistant helping researchers study persuasion. 
Your task is to paraphrase provided texts so that the meaning, and original persuasion techniques are strictly preserved.
Do not make the provided texts less or more persuasive. Keep the same level of persuasion. 
Definition of persuasion is as follows: Persuasive text is characterized by a specific use of language in order to influence readers.
We distinguish the following high-level persuasion approaches: Attack on reputation, Justification, Simplification, Distraction, Call, Manipulative wording.
You will receive the original passage. Then, paraphrase the passage, ensuring the paraphrased version contains the exact same level of persuasion and meaning. Your final output should only be the paraphrased text.\\
\textbf{User Prompt:} \\
Only paraphrase the following passage in \textit{<language>}\\
The passage is: \textit{<Human-written, persuasive text to paraphrase>} \\
Keep the same level of persuasion.
Provide the paraphrased text in 
\textit{<language>}, in valid JSON format: \\
\texttt{\{} \\
\hspace*{1.5em}\texttt{"generated\_text":"Your paraphrased text in \textit{<language>} here."} \\
\texttt{\}} \\

\medskip
  \end{tcolorbox}
  \caption{Prompt template used for persuasive texts generation with LLMs using the \textit{Paraphrasing Generation} approach.}
  \label{fig:prompt_paraphrased}
\end{figure*}

\begin{figure*}[ht]
  \centering
    \begin{tcolorbox}[colback=Orange!20!white, colframe=DarkOrange!70!black, 
    width=\textwidth, boxrule=0.5pt, arc=3pt, auto outer arc,
    fonttitle=\bfseries, title=Generation approach: Rewriting Generation - Subtle Persuasion prompt]
\scriptsize
\textbf{System Prompt:} \\
You are an assistant helping researchers study persuasion. 
Your task is to rewrite provided texts so that the persuasion is more subtle and harder to detect, while strictly preserving all original persuasion techniques.
Definition of persuasion is as follows: Persuasive text is characterized by a specific use of language in order to influence readers.
We distinguish the following high-level persuasion approaches: Attack on reputation, Justification, Simplification, Distraction, Call, Manipulative wording.
You will receive the original passage. Then, rewrite the passage, ensuring the rewritten version contains the exact meaning. Your final output should only be the rewritten text.\\
\textbf{User Prompt:} \\
Only rewrite the following passage in \textit{<language>} \textbf{so that the persuasion is more subtle and harder to detect.}\\
The passage is: \textit{<Human-written, persuasive text to rewrite>} \\
Provide the rewritten text in \textit{<language>}, in valid JSON format: \\
\texttt{\{} \\
\hspace*{1.5em}\texttt{"generated\_text":"Your rewritten text in \textit{<language>} here."} \\
\texttt{\}} \\

\medskip
  \end{tcolorbox}
  \caption{Prompt template used for persuasive texts generation with LLMs using the \textit{Rewriting Generation - Subtle Persuasion} approach.}
  \label{fig:prompt_subtle}
\end{figure*}

\begin{figure*}[ht]
  \centering
\begin{tcolorbox}[colback=orange!5!white, colframe=orange!50!black, 
    width=\textwidth, boxrule=0.5pt, arc=3pt, auto outer arc,
    fonttitle=\bfseries, title=Generation approach: Rewriting Generation - Intensive Persuasion prompt]
\scriptsize
\textbf{System Prompt:} \\
                  You are an assistant helping researchers study persuasion. 
                Your task is to rewrite provided texts so that the persuasion is refined for stronger persuasive effect, while strictly preserving all original persuasion techniques.
                Definition of persuasion is as follows: Persuasive text is characterized by a specific use of language in order to influence readers.
                We distinguish the following high-level persuasion approaches: Attack on reputation, Justification, Simplification, Distraction, Call, Manipulative wording.
                You will receive the original passage. Then, rewrite the passage, ensuring the rewritten version contains the exact meaning. Your final output should only be the rewritten text. \\
\textbf{User Prompt:} \\
Only rewrite the following passage in \textit{<language>} \textbf{so that the persuasion is refined for stronger persuasive effect.}\\
The passage is: \textit{<Human-written, persuasive text to paraphrase>} \\
Provide the rewritten text in \textit{<language>}, in valid JSON format: \\
\texttt{\{} \\
\hspace*{1.5em}\texttt{"generated\_text":"Your rewritten text in \textit{<language>} here."} \\
\texttt{\}} \\

\medskip
  \end{tcolorbox}
  \caption{Prompt template used for persuasive texts generation with LLMs using the \textit{Rewriting Generation - Intensive Persuasion} approach.}
  \label{fig:prompt_intensified}
\end{figure*}

\begin{figure*}[ht]
  \centering
  \begin{tcolorbox}[colback=blue!10!white, colframe=blue!40!black, 
    width=\textwidth, boxrule=0.5pt, arc=3pt, auto outer arc,
    fonttitle=\bfseries, title=Generation approach: Open-ended Generation prompt]
\scriptsize
\textbf{System Prompt:} \\
You are an assistant helping researchers study persuasion. 
Your task is to generate a short text based on a provided passage. The short text must contain persuasion.
Definition of persuasion is as follows: Persuasive text is characterized by a specific use of language in order to influence readers.
We distinguish the following high-level persuasion approaches: Attack on reputation, Justification, Simplification, Distraction, Call, Manipulative wording.
Your final output should only be the generated text.\\
\textbf{User Prompt:} \\
Generate a text in \textit{<language>} based on the following passage in \textit{<language>}.\\
The passage is: \textcolor{blue!50!gray!80!black}{\textit{<Summarized, factual, and non-persuasive input sentence>}} \\
The generated text must contain persuasion. Provide the generated text in valid JSON format:\\
\texttt{\{} \\
\hspace*{1.5em}\texttt{"generated\_text":"Your generated text in \textit{<language>} here."} \\
\texttt{\}} \\

\medskip
  \end{tcolorbox}
  
\begin{tcolorbox}[colback=blue!20!gray!15!white, colframe=blue!50!gray!80!black, 
    width=\textwidth, boxrule=0.5pt, arc=3pt, auto outer arc,
    fonttitle=\bfseries, title=Obtaining a summarized\, factual\ input sentence prompt]
\scriptsize
\textbf{System Prompt:} \\
You are a journalist assistant. 
Your task is to convert the provided text passage into a direct, single-sentence text.
Do not add context such as 'The speaker said...', 'The passage is about...', 'The statement suggests...', etc. 
Keep the meaning intact but make it stand alone. 
Do not add any additional information or actors.\\
\textbf{User Prompt:} \\
Restate the following passage in \textit{<language>} as a single-sentence, neutral text in \textit{<language>}.\\
The passage is: \textit{<Human-written, persuasive text to summarize>} \\
Return in valid JSON format:\\
\texttt{\{} \\
\hspace*{1.5em}\texttt{"generated\_text":"Your restated sentence in  \textit{<language>} here."} \\
\texttt{\}} \\

\medskip
  \end{tcolorbox}
  \caption{Prompt template used for persuasive texts generation with LLMs using the \textit{Open-ended Generation} approach along with the prompt template used to obtain a summarized, factual, and non-persuasive sentence input from human-written persuasive texts.}
  \label{fig:open_ended}
\end{figure*}

\begin{figure*}[ht]
  \centering
  \begin{tcolorbox}[colback=Crimson!20!white, colframe=Crimson!60!black, 
    width=\textwidth, boxrule=0.5pt, arc=3pt, auto outer arc,
    fonttitle=\bfseries, title=Binary Detection of Persuasion prompt]
\scriptsize
\textbf{System Prompt:} \\
                 You are an assistant who detects persuasion in text. Persuasive text is characterized by a specific use of language in order to influence readers. 
                We distinguish the following high-level persuasion approaches: Attack on reputation, Justification, Simplification, Distraction, Call, Manipulative wording.
                You are the expert who detects high-level persuasion. \\
\textbf{User Prompt:} \\
Analyze the following passage: \textit{<Text to analyze>} \\
Decide if the passage contains persuasion. You are very conservative in your final decisions and when you are not fully sure you answer 'No'. Do not provide any additional text, just JSON. Give only your final answer 'Yes' or 'No' in valid JSON format:\\
\texttt{\{} \\
\hspace*{1.5em}\texttt{"decision"}: \texttt{"'Yes' if passage contains persuasion, 'No' otherwise."} \\
\texttt{\}} \\

\medskip
  \end{tcolorbox}
  \caption{Prompt template used for binary classification of persuasive texts with LLMs.}
  \label{fig:prompt_binary_llms}
\end{figure*}

\section{Details on LLMs used in Experiments and selection rationale.}
\label{app:llm_used_selection_rationale}

In our experiments, we employed four state-of-the-art LLMs: \textit{GPT-4.1 Mini}, \textit{Gemini 2.0 Flash}, \textit{Gemma 3 27B-IT}, and \textit{Llama 3.3 70B}. Our selection aimed to cover widely recognized, high-performing models while balancing accessibility and cost. Additionally, we included two open-weight models to provide experiments that can be reproduced without reliance on closed API-based models. Table \ref{tab:llm-models} summarizes the LLMs used, including their knowledge cutoff dates, access methods, licenses, and model sizes.

\begin{table*}[ht!]
\scriptsize
\centering
\begin{tabular}{lllll}
\toprule
 \textbf{API Model Name} & \textbf{Knowledge Cutoff Date} & \textbf{Access Details} & \textbf{License} & \textbf{Model Size}\\\midrule
 \texttt{gemini-2.0-flash} & June 2024 & Google API 07.2025 & Commercial & Not Disclosed\\ 
 \texttt{gpt-4.1-mini-2025-04-14} & June 2024 & OpenAI API 07.2025 & Commercial & Not Disclosed\\ 
 \texttt{meta-llama/Llama-3.3-70B-Instruct} &  December 2023 & DeepInfra API 07.2025 & Meta Llama 3 Community & 70B \\ 
 \texttt{google/gemma-3-27b-it} & August 2024 & DeepInfra API 07.2025 & Gemma Terms of Use & 27B \\ 
\bottomrule
\end{tabular}
\caption{Large Language Models used in our experiments.}
\label{tab:llm-models}
\end{table*}

\section{Annotation Guidelines for Dataset Evaluation}
\label{app:annotation_guidelines}

\subsection{Annotation Guidelines for Sentences Verification}
\label{app:pre_annotation_guidelines}

\paragraph{Purpose of the Annotation Task.} The goal of this annotation task is to evaluate whether each LLM-generated \brownehl{<sentence>} accurately reflects content present in its corresponding source human text. Annotators must independently judge whether the \brownehl{<sentence>} faithfully reflects information explicitly stated in the source text, without adding, or altering factual content.

\paragraph{General Annotation Procedure.}
\begin{enumerate} [nosep, leftmargin=*]
    \item Read the source persuasive human text in full to understand its factual content and context.
    \item Read the generated \brownehl{<sentence>} carefully and evaluate it against the factual correspondence.
    \item Assign one binary label: \\ Factual? Yes (1) / No (0)
    \item Do not consider any stylistic preferences, or grammar.
\end{enumerate}

Annotators should make decisions independently, without discussing individual cases during the evaluation phase.

\paragraph{Factual Correspondence Annotation.}
\begin{enumerate}[nosep, leftmargin=*]
    \item All information in the \brownehl{<sentence>} is explicitly stated in the source text.
    \begin{itemize}
        \item No invented facts.
        \item The \brownehl{<sentence>} does not introduce generalizations (e.g., Fact present in a source text: "Adam Smith fainted after COVID-19 vaccination" $\rightarrow$ invalid \brownehl{<sentence>}: "People fainted after COVID-19 vaccination")
        \item No added assumptions or interpretations.
    \end{itemize}
    \item No main factual information from the source text is omitted in a way that distorts meaning.
    \item The \brownehl{<sentence>} is neutral and descriptive - Its purpose must be to summarize factual content, not to evaluate, interpret, or advise.
    \item Statements must be verifiable based solely on the source text. Annotators should not use outside knowledge.
\end{enumerate}

Examples of factual errors (should be labeled "No"):
\begin{itemize}[nosep, leftmargin=*]
    \item Adding additional events or statistics not in the source
    \item Reframing a claim as a fact (e.g., converting someone’s opinion into an asserted truth)
    \item Omitting a main fact presented in source text that changes meaning.
\end{itemize}

\subsection{Annotation Guidelines for Persuaficial Dataset Evaluation}
\label{app:post_annotation_guidelines}

These guidelines describe the annotation protocol for evaluating LLM-generated persuasive texts in the Persuaficial dataset. Each generated text is independently annotated by two annotators. The post-generation evaluation focuses on three key dimensions:
\begin{itemize}[nosep, leftmargin=*]
    \item Factual Correspondence: Is the generated text faithful to the target factual content?
    \item Persuasiveness: Does the text contain genuine persuasion?
    \item Instruction Adherence: Does the text follow the specific persuasion instruction for its generation approach?
\end{itemize}

\paragraph{Factual Correspondence.} Goal of this step is to ensure the generated text accurately reflects the source content.

Instructions:
\begin{itemize}[nosep, leftmargin=*]
    \item Open-Ended Generation: Refer to the factual sentence.
    \item Paraphrasing / Rewriting Approaches: Refer to the original passage.
\end{itemize}

Assessment:
\begin{itemize}[nosep, leftmargin=*]
    \item Valid (represented as 1): Text preserves the factual meaning of the source without introducing errors or contradictions.
    \item Invalid (represented as 0): Text contains factual inaccuracies, omissions, or misrepresentations.
\end{itemize}
Note: Only factual distortion triggers an Invalid label.

\paragraph{Persuasiveness}
The generated text must contain any persuasive elements.

For this task, we define persuasive text as text characterized by a specific use of language in order to influence readers \cite{piskorski2023news, piskorski2023multilingual}. The generated text must be labeled as persuasive (represented as 1) if it exhibits any of the following high-level persuasion strategies:
\begin{itemize}[nosep, leftmargin=*]
    \item Attack on reputation: the argument does not address the topic itself, but targets the participant (personality, experience, deeds, etc.) in order to question and/or to undermine his credibility. The object of the argumentation can also refer to a group of individuals, an organization, an object, or an activity,
    \item Justification: the argument is made of two parts, a statement and an explanation or appeal, where the latter is used to justify and/or to support the statement,
    \item Simplification: the argument excessively simplifies a problem, usually regarding the cause, the consequence, or the existence of choices,
    \item Distraction: the argument takes focus away from the main topic or argument to distract the reader,
    \item Call: the text is not an argument but an encouragement to act or to think in a particular way,
    \item Manipulative wording: the text is not an argument per se, but uses specific language, which contains words or phrases that are either non-neutral, confusing, exaggerating, loaded, etc., in order to impact the reader emotionally.
\end{itemize}

If any of these strategies are present, the sentence must be labeled 1 (persuasive) for the persuasiveness criterion.

\paragraph{Instruction Adherence.} The goal is to verify that the text aligns with the intended generation approach.

Instructions for Annotators:
\begin{enumerate}[nosep, leftmargin=*]
    \item Compare the generated text to the prompt provided to the model.
    \item Label Compliant (represented as 1) if the text follows the prompt goal; Non-Compliant (represented as 0) if it deviates.
\end{enumerate}

\section{Persuaficial Dataset - Additional Statistics}
\label{appendix:add_persuaficial_stats}

Table~\ref{tab:all_basic_stats} summarizes the basic statistics of both human-written and LLM-generated texts in the Persuaficial dataset. The table reports average word, average characters and number of words across all languages and generation types. Moreover, we show statistics in general for full Persuaficial dataset.

\begin{table}
\centering
\scriptsize
\setlength{\tabcolsep}{3pt}
\renewcommand{\arraystretch}{1.3} 
\begin{tabular}{llrrr}
\toprule
 &  & $Avg_w$ & $Avg_{ch}$ & Count \\
dataset & type &  &  &  \\
\midrule
\multirow[t]{5}{*}{Persuaficial} & Rewriting (intensified) & 87 & 605 & 16320 \\
 & Open-ended & 65 & 450 & 16320 \\
 & Paraphrasing & 81 & 538 & 16320 \\
 & Rewriting (subtle) & 87 & 568 & 16320 \\
\cline{1-5}
\multirow[t]{5}{*}{English} & Human & 117 & 695 & 3000 \\
 & Rewriting (intensified) & 118 & 791 & 6000 \\
 & Open-ended & 60 & 391 & 6000 \\
 & Paraphrasing & 111 & 727 & 6000 \\
 & Rewriting (subtle) & 110 & 742 & 6000 \\
\cline{1-5}
\multirow[t]{5}{*}{French} & Human & 46 & 288 & 1000 \\
 & Rewriting (intensified) & 75 & 498 & 2000 \\
 & Open-ended & 76 & 498 & 2000 \\
 & Paraphrasing & 66 & 430 & 2000 \\
 & Rewriting (subtle) & 69 & 465 & 2000 \\
\cline{1-5}
\multirow[t]{5}{*}{German} & Human & 44 & 314 & 1000 \\
 & Rewriting (intensified) & 68 & 512 & 2320 \\
 & Open-ended & 71 & 524 & 2320 \\
 & Paraphrasing & 59 & 434 & 2320 \\
 & Rewriting (subtle) & 63 & 476 & 2320 \\
\cline{1-5}
\multirow[t]{5}{*}{Italian} & Human & 48 & 313 & 1000 \\
 & Rewriting (intensified) & 75 & 511 & 2000 \\
 & Open-ended & 75 & 507 & 2000 \\
 & Paraphrasing & 66 & 443 & 2000 \\
 & Rewriting (subtle) & 70 & 488 & 2000 \\
\cline{1-5}
\multirow[t]{5}{*}{Polish} & Human & 46 & 327 & 1000 \\
 & Rewriting (intensified) & 70 & 526 & 2000 \\
 & Open-ended & 61 & 454 & 2000 \\
 & Paraphrasing & 62 & 459 & 2000 \\
 & Rewriting (subtle) & 66 & 493 & 2000 \\
\cline{1-5}
\multirow[t]{5}{*}{Russian} & Human & 40 & 285 & 1000 \\
 & Rewriting (intensified) & 55 & 417 & 2000 \\
 & Open-ended & 58 & 442 & 2000 \\
 & Paraphrasing & 48 & 360 & 2000 \\
 & Rewriting (subtle) & 52 & 400 & 2000 \\
\cline{1-5}
\bottomrule
\end{tabular}
\caption{Basic statistics for human-written and for LLM-generated persuasive texts in our Persuaficial dataset. We present basic statistics for general full dataset, but also on samples that present all languages. $Avg_{w}$ stands for average words and $Avg_{ch}$ stands for average characters.}
\label{tab:all_basic_stats}
\end{table}

\section{Persuasion Detection - Experimental Setup Details}
\label{appendix:cls_details}
\subsection{Evaluation Text Sample Creation}

Our evaluation framework comprised a total of 68 distinct classification experiments for each of the eight source datasets (eight as SemEval data can be counted as 6 datasets each in different language). This setup involved testing every combination of four generation approaches and four generating LLMs. The resulting 16 sets of machine-generated text, along with a baseline of human-written persuasive text, were then evaluated by four different classifying LLMs, leading to the (16 + 1) × 4 = 68 experimental conditions.\\

Our evaluation framework is designed to isolate the impact of these generated texts.
Each experiment's evaluation set is composed of two halves:
\begin{itemize}[nosep, leftmargin=*]
    \item A constant set of 500 human-written, non-persuasive texts from the original dataset. Predictions for this set were calculated once for each model and reused across all experiments for that dataset.
    \item A variable set of 500 persuasive texts, which consists of the LLM-generated samples for a given generation approach.
\end{itemize}
Consequently, any variation in the F\textsubscript{1} score between different generation models on the same dataset is attributable solely to the model's performance on the generated persuasive samples. For our baseline experiments, labeled 'Human-written', these persuasive samples are the original human-written texts from the dataset.

\subsection{Experimental Setup for LLM Classification Process}
\label{app:classification_llm_exp_setup}
We employed four Large Language Models.
The open models include \textit{Gemma~3~27b~it} and \textit{Llama~3.3~70B}. The commercial models are \textit{Gemini~2.0~Flash} and \textit{GPT~4.1~Mini}.
To ensure determinism in the classification predictions, we set the classification temperature to 0.0. Rationale for models selection provided in Appendix \ref{app:llm_used_selection_rationale}.

\subsection{Prompt Templates for Persuasion Detection}
\label{app:prompt_templates_persuasion_detection}
Figure \ref{fig:prompt_binary_llms} shows a prompt template used during the LLM persuasion detection process. In our prompts, we adopt the concise definition of persuasion proposed by \citet{piskorski2023multilingual, piskorski2023news}: “\textit{Persuasive text is characterized by a specific use of language in order to influence the reader}”.



\section{Human vs. AI-generated Persuasion Detection  - Detailed Results}
\label{app:detailed_results_ai_human_persuasion_detection}

\subsection{Detailed results on English texts of Persuaficial dataset}
\label{app:human_machine_persuasion_text_class_results_english}

In this appendix, we present detailed F\textsubscript{1} scores for persuasion detection across different subsets of the Persuaficial datasets and their human-written counterparts. Results are presented for each LLM generation strategy. 
Table~\ref{tab:semeval_english_results} reports results for SemEval 2023 Task 3 texts and their AI-generated counterparts, Table~\ref{tab:dipromats_results} for the DIPROMATS 2024 dataset, and 
Table~\ref{tab:winning_arguments_results} for the ChangeMyView dataset. For each dataset, classifier performance on human-written texts (first column) is compared with performance on LLM-generated texts produced via paraphrasing, rewriting with subtle or intensive persuasion, and open-ended generation. 
Results are further broken down by both the generating model and the classifier model, highlighting how different generation approaches influence the detectability of persuasion.

\begin{table*}[!ht]
\centering
\scriptsize
\begin{tabular}{lccccc}
\hline
\textbf{Classifier Models} & 
\textbf{Human-written} & \textbf{Paraphrasing Generation} & 
\multicolumn{2}{c}{\textbf{Rewriting Generation}} & 
\textbf{Open-ended Generation} \\ 
 & & & \textbf{Subtle Persuasion} & \textbf{Intensive Persuasion} & \\ \hline

\multicolumn{6}{l}{\textit{\textbf{Generating model: GPT 4.1 Mini}}} \\
GPT 4.1 Mini & 0.7398 & \perc{0.7398}{0.6984} & \perc{0.7398}{0.3837} & \perc{0.7398}{0.7638} & \perc{0.7398}{0.8969} \\
Llama 3.3 70B & 0.7459 & \perc{0.7459}{0.7310} & \perc{0.7459}{0.4444} & \perc{0.7459}{0.7757} & \perc{0.7459}{0.8741} \\
Gemma 3 27b it & 0.7572 & \perc{0.7572}{0.7561} & \perc{0.7572}{0.6213} & \perc{0.7572}{0.8007} & \perc{0.7572}{0.8562} \\
Gemini 2.0 Flash & 0.7551 & \perc{0.7551}{0.7487} & \perc{0.7551}{0.6407} & \perc{0.7551}{0.7780} & \perc{0.7551}{0.8117} \\
\hline
\multicolumn{6}{l}{\textit{\textbf{Generating model: Llama 3.3 70B}}} \\
GPT 4.1 Mini & 0.7398 & \perc{0.7398}{0.6831} & \perc{0.7398}{0.4411} & \perc{0.7398}{0.7870} & \perc{0.7398}{0.8969} \\
Llama 3.3 70B & 0.7459 & \perc{0.7459}{0.6911} & \perc{0.7459}{0.4811} & \perc{0.7459}{0.7791} & \perc{0.7459}{0.8741} \\
Gemma 3 27b it & 0.7572 & \perc{0.7572}{0.7479} & \perc{0.7572}{0.6746} & \perc{0.7572}{0.8082} & \perc{0.7572}{0.8562} \\
Gemini 2.0 Flash & 0.7551 & \perc{0.7551}{0.7476} & \perc{0.7551}{0.6691} & \perc{0.7551}{0.7921} & \perc{0.7551}{0.8117} \\
\hline
\multicolumn{6}{l}{\textit{\textbf{Generating model: Gemma 3 27b it}}} \\
GPT 4.1 Mini & 0.7398 & \perc{0.7398}{0.7025} & \perc{0.7398}{0.3445} & \perc{0.7398}{0.8611} & \perc{0.7398}{0.8969} \\
Llama 3.3 70B & 0.7459 & \perc{0.7459}{0.7222} & \perc{0.7459}{0.4118} & \perc{0.7459}{0.8469} & \perc{0.7459}{0.8741} \\
Gemma 3 27b it & 0.7572 & \perc{0.7572}{0.7675} & \perc{0.7572}{0.6241} & \perc{0.7572}{0.8383} & \perc{0.7572}{0.8562} \\
Gemini 2.0 Flash & 0.7551 & \perc{0.7551}{0.7614} & \perc{0.7551}{0.6202} & \perc{0.7551}{0.8039} & \perc{0.7551}{0.8117} \\
\hline
\multicolumn{6}{l}{\textit{\textbf{Generating model: Gemini 2.0 Flash}}} \\
GPT 4.1 Mini & 0.7398 & \perc{0.7398}{0.7188} & \perc{0.7398}{0.4430} & \perc{0.7398}{0.8472} & \perc{0.7398}{0.8949} \\
Llama 3.3 70B & 0.7459 & \perc{0.7459}{0.7385} & \perc{0.7459}{0.4936} & \perc{0.7459}{0.8428} & \perc{0.7459}{0.8741} \\
Gemma 3 27b it & 0.7572 & \perc{0.7572}{0.7652} & \perc{0.7572}{0.6613} & \perc{0.7572}{0.8362} & \perc{0.7572}{0.8562} \\
Gemini 2.0 Flash & 0.7551 & \perc{0.7551}{0.7583} & \perc{0.7551}{0.6787} & \perc{0.7551}{0.8059} & \perc{0.7551}{0.8117} \\
\hline
\end{tabular}
\caption{F\textsubscript{1} scores for persuasion detection on English data sample of Persuaficial. More specifically, on sample of Persuaficial generated using English texts from SemEval 2023 Task 3 dataset. The first column reports performance on English texts from SemEval 2023 Task 3 human-annotated texts. The remaining columns show performance on LLM-generated English counterparts.}
\label{tab:semeval_english_results}
\end{table*}

\begin{table*}[!ht]
\centering
\scriptsize
\begin{tabular}{lccccc}
\hline
\textbf{Classifier Models} & 
\textbf{Human-written} & \textbf{Paraphrasing Generation} & 
\multicolumn{2}{c}{\textbf{Rewriting Generation}} & 
\textbf{Open-ended Generation} \\ 
 & & & \textbf{Subtle Persuasion} & \textbf{Intensive Persuasion} & \\ \hline

\multicolumn{6}{l}{\textit{\textbf{Generating model: GPT 4.1 Mini}}} \\
GPT 4.1 Mini & 0.7567 & \perc{0.7567}{0.7435} & \perc{0.7567}{0.4948} & \perc{0.7567}{0.7866} & \perc{0.7567}{0.8666} \\
Llama 3.3 70B         & 0.7471 & \perc{0.7471}{0.7348} & \perc{0.7471}{0.5795} & \perc{0.7471}{0.7679} & \perc{0.7471}{0.8292} \\
Gemma 3 27b it         & 0.7473 & \perc{0.7473}{0.7441} & \perc{0.7473}{0.6308} & \perc{0.7473}{0.7607} & \perc{0.7473}{0.7994} \\
Gemini 2.0 Flash        & 0.7518 & \perc{0.7518}{0.7449} & \perc{0.7518}{0.6711} & \perc{0.7518}{0.7595} & \perc{0.7518}{0.7680} \\
\hline
\multicolumn{6}{l}{\textit{\textbf{Generating model: Llama 3.3 70B}}} \\
GPT 4.1 Mini & 0.7567 & \perc{0.7567}{0.7338} & \perc{0.7567}{0.5595} & \perc{0.7567}{0.7967} & \perc{0.7567}{0.8666} \\
Llama 3.3 70B         & 0.7471 & \perc{0.7471}{0.7314} & \perc{0.7471}{0.6251} & \perc{0.7471}{0.7775} & \perc{0.7471}{0.8292} \\
Gemma 3 27b it         & 0.7473 & \perc{0.7473}{0.7410} & \perc{0.7473}{0.6928} & \perc{0.7473}{0.7749} & \perc{0.7473}{0.7994} \\
Gemini 2.0 Flash        & 0.7518 & \perc{0.7518}{0.7400} & \perc{0.7518}{0.7002} & \perc{0.7518}{0.7652} & \perc{0.7518}{0.7680} \\
\hline
\multicolumn{6}{l}{\textit{\textbf{Generating model: Gemma 3 27b it}}} \\
GPT 4.1 Mini & 0.7567 & \perc{0.7567}{0.7672} & \perc{0.7567}{0.3951} & \perc{0.7567}{0.8404} & \perc{0.7567}{0.8666} \\
Llama 3.3 70B         & 0.7471 & \perc{0.7471}{0.7449} & \perc{0.7471}{0.4898} & \perc{0.7471}{0.8074} & \perc{0.7471}{0.8292} \\
Gemma 3 27b it         & 0.7473 & \perc{0.7473}{0.7504} & \perc{0.7473}{0.5777} & \perc{0.7473}{0.7936} & \perc{0.7473}{0.7994} \\
Gemini 2.0 Flash        & 0.7518 & \perc{0.7518}{0.7459} & \perc{0.7518}{0.6232} & \perc{0.7518}{0.7662} & \perc{0.7518}{0.7680} \\
\hline
\multicolumn{6}{l}{\textit{\textbf{Generating model: Gemini 2.0 Flash}}} \\
GPT 4.1 Mini & 0.7567 & \perc{0.7567}{0.7399} & \perc{0.7567}{0.5353} & \perc{0.7567}{0.8163} & \perc{0.7567}{0.8666} \\
Llama 3.3 70B         & 0.7471 & \perc{0.7471}{0.7336} & \perc{0.7471}{0.5838} & \perc{0.7471}{0.7911} & \perc{0.7471}{0.8292} \\
Gemma 3 27b it         & 0.7473 & \perc{0.7473}{0.7483} & \perc{0.7473}{0.6383} & \perc{0.7473}{0.7838} & \perc{0.7473}{0.7994} \\
Gemini 2.0 Flash        & 0.7518 & \perc{0.7518}{0.7400} & \perc{0.7518}{0.6711} & \perc{0.7518}{0.7652} & \perc{0.7518}{0.7680} \\

\hline
\end{tabular}
\caption{F\textsubscript{1} scores for persuasion detection on English data sample of Persuaficial. More specifically, on sample of Persuaficial generated using DIPROMATS 2024 dataset. The first column reports performance on DIPROMATS 2024 human-annotated texts. The remaining columns show performance on LLM-generated counterparts.}
\label{tab:dipromats_results}
\end{table*}

\begin{table*}[!ht]
\centering
\scriptsize
\begin{tabular}{lccccc}
\hline
\textbf{Classifier Models} & 
\textbf{Human-written} & \textbf{Paraphrasing Generation} & 
\multicolumn{2}{c}{\textbf{Rewriting Generation}} & 
\textbf{Open-ended Generation} \\ 
 & & & \textbf{Subtle Persuasion} & \textbf{Intensive Persuasion} & \\ \hline

\multicolumn{6}{l}{\textit{\textbf{Generating model: GPT 4.1 Mini}}} \\
GPT 4.1 Mini & 0.6233 & \perc{0.6233}{0.6337} & \perc{0.6233}{0.4941} & \perc{0.6233}{0.6582} & \perc{0.6233}{0.7148} \\
Llama 3.3 70B         & 0.6517 & \perc{0.6517}{0.6546} & \perc{0.6517}{0.5665} & \perc{0.6517}{0.6667} & \perc{0.6517}{0.6831} \\
Gemma 3 27b it         & 0.6644 & \perc{0.6644}{0.6745} & \perc{0.6644}{0.6388} & \perc{0.6644}{0.6809} & \perc{0.6644}{0.6836} \\
Gemini 2.0 Flash        & 0.6671 & \perc{0.6671}{0.6662} & \perc{0.6671}{0.6431} & \perc{0.6671}{0.6726} & \perc{0.6671}{0.6770} \\
\hline
\multicolumn{6}{l}{\textit{\textbf{Generating model: Llama 3.3 70B}}} \\
GPT 4.1 Mini & 0.6233 & \perc{0.6233}{0.6137} & \perc{0.6233}{0.4619} & \perc{0.6233}{0.6481} & \perc{0.6233}{0.7148} \\
Llama 3.3 70B         & 0.6517 & \perc{0.6517}{0.6307} & \perc{0.6517}{0.5226} & \perc{0.6517}{0.6564} & \perc{0.6517}{0.6831} \\
Gemma 3 27b it         & 0.6644 & \perc{0.6644}{0.6644} & \perc{0.6644}{0.6133} & \perc{0.6644}{0.6772} & \perc{0.6644}{0.6845} \\
Gemini 2.0 Flash        & 0.6671 & \perc{0.6671}{0.6607} & \perc{0.6671}{0.6112} & \perc{0.6671}{0.6717} & \perc{0.6671}{0.6770} \\
\hline
\multicolumn{6}{l}{\textit{\textbf{Generating model: Gemma 3 27b it}}} \\
GPT 4.1 Mini & 0.6233 & \perc{0.6233}{0.6572} & \perc{0.6233}{0.4840} & \perc{0.6233}{0.7009} & \perc{0.6233}{0.7148} \\
Llama 3.3 70B         & 0.6517 & \perc{0.6517}{0.6574} & \perc{0.6517}{0.5515} & \perc{0.6517}{0.6758} & \perc{0.6517}{0.6831} \\
Gemma 3 27b it         & 0.6644 & \perc{0.6644}{0.6754} & \perc{0.6644}{0.6349} & \perc{0.6644}{0.6836} & \perc{0.6644}{0.6845} \\
Gemini 2.0 Flash        & 0.6671 & \perc{0.6671}{0.6689} & \perc{0.6671}{0.6259} & \perc{0.6671}{0.6762} & \perc{0.6671}{0.6770} \\
\hline
\multicolumn{6}{l}{\textit{\textbf{Generating model: Gemini 2.0 Flash}}} \\
GPT 4.1 Mini & 0.6233 & \perc{0.6233}{0.6379} & \perc{0.6233}{0.5226} & \perc{0.6233}{0.6885} & \perc{0.6233}{0.7148} \\
Llama 3.3 70B         & 0.6517 & \perc{0.6517}{0.6527} & \perc{0.6517}{0.5740} & \perc{0.6517}{0.6776} & \perc{0.6517}{0.6831} \\
Gemma 3 27b it         & 0.6644 & \perc{0.6644}{0.6690} & \perc{0.6644}{0.6465} & \perc{0.6644}{0.6818} & \perc{0.6644}{0.6845} \\
Gemini 2.0 Flash        & 0.6671 & \perc{0.6671}{0.6689} & \perc{0.6671}{0.6374} & \perc{0.6671}{0.6753} & \perc{0.6671}{0.6770} \\

\hline
\end{tabular}
\caption{F\textsubscript{1} scores for persuasion detection on English data sample of Persuaficial. More specifically, on sample of Persuaficial generated using ChangeMyView dataset. The first column reports performance on ChangeMyView human-annotated texts. The remaining columns show performance on LLM-generated counterparts.}
\label{tab:winning_arguments_results}
\end{table*}

\subsection{Detailed results on non-English texts of Persuaficial dataset}
\label{app:human_machine_persuasion_text_class_results_non_english}

In addition to the English results, we provide detailed F\textsubscript{1} scores for persuasion detection on other language-specific subsets of the Persuaficial datasets. 
Table~\ref{tab:semeval_german_results} reports results for German texts from SemEval 2023 Task 3 and their LLM-generated counterparts, Table~\ref{tab:semeval_french_results} for French texts, Table~\ref{tab:semeval_italian_results} for Italian texts, Table~\ref{tab:semeval_polish_results} for Polish texts, and Table~\ref{tab:semeval_russian_results} for Russian texts. 
For each dataset, classifier performance on human-written texts (first column) is compared with performance on LLM-generated texts produced via paraphrasing, rewriting with subtle or intensive persuasion, and open-ended generation. The results are further broken down by both the generating model and the classifier model, demonstrating how different generation approaches influence the detectability of persuasion across languages.

\begin{table*}[!ht]
\centering
\scriptsize
\begin{tabular}{lccccc}
\hline
\textbf{Classifier Models} & 
\textbf{Human-written} & \textbf{Paraphrasing Generation} & 
\multicolumn{2}{c}{\textbf{Rewriting Generation}} & 
\textbf{Open-ended Generation} \\ 
 & & & \textbf{Subtle Persuasion} & \textbf{Intensive Persuasion} & \\ \hline

\multicolumn{6}{l}{\textit{\textbf{Generating model: GPT 4.1 Mini}}} \\
GPT 4.1 Mini & 0.7203 & \perc{0.7203}{0.6999} & \perc{0.7203}{0.4677} & \perc{0.7203}{0.7959} & \perc{0.7203}{0.9416} \\
Llama 3.3 70B & 0.7361 & \perc{0.7361}{0.7239} & \perc{0.7361}{0.4574} & \perc{0.7361}{0.8004} & \perc{0.7361}{0.9347} \\
Gemma 3 27b it & 0.7655 & \perc{0.7655}{0.7644} & \perc{0.7655}{0.6553} & \perc{0.7655}{0.8220} & \perc{0.7655}{0.9006} \\
Gemini 2.0 Flash & 0.7903 & \perc{0.7903}{0.7768} & \perc{0.7903}{0.6871} & \perc{0.7903}{0.8244} & \perc{0.7903}{0.8593} \\
\hline
\multicolumn{6}{l}{\textit{\textbf{Generating model: Llama 3.3 70B}}} \\
GPT 4.1 Mini & 0.7203 & \perc{0.7203}{0.7102} & \perc{0.7203}{0.4604} & \perc{0.7203}{0.7836} & \perc{0.7203}{0.9416} \\
Llama 3.3 70B & 0.7361 & \perc{0.7361}{0.7128} & \perc{0.7361}{0.4681} & \perc{0.7361}{0.7971} & \perc{0.7361}{0.9347} \\
Gemma 3 27b it & 0.7655 & \perc{0.7655}{0.7666} & \perc{0.7655}{0.6901} & \perc{0.7655}{0.8200} & \perc{0.7655}{0.9006} \\
Gemini 2.0 Flash & 0.7903 & \perc{0.7903}{0.7816} & \perc{0.7903}{0.7156} & \perc{0.7903}{0.8298} & \perc{0.7903}{0.8593} \\
\hline
\multicolumn{6}{l}{\textit{\textbf{Generating model: Gemma 3 27b it}}} \\
GPT 4.1 Mini & 0.7203 & \perc{0.7203}{0.7303} & \perc{0.7203}{0.3881} & \perc{0.7203}{0.8996} & \perc{0.7203}{0.9407} \\
Llama 3.3 70B & 0.7361 & \perc{0.7361}{0.7252} & \perc{0.7361}{0.3779} & \perc{0.7361}{0.8946} & \perc{0.7361}{0.9339} \\
Gemma 3 27b it & 0.7655 & \perc{0.7655}{0.7877} & \perc{0.7655}{0.6411} & \perc{0.7655}{0.8824} & \perc{0.7655}{0.8998} \\
Gemini 2.0 Flash & 0.7903 & \perc{0.7903}{0.7978} & \perc{0.7903}{0.6644} & \perc{0.7903}{0.8499} & \perc{0.7903}{0.8584} \\
\hline
\multicolumn{6}{l}{\textit{\textbf{Generating model: Gemini 2.0 Flash}}} \\
GPT 4.1 Mini & 0.7203 & \perc{0.7203}{0.7425} & \perc{0.7203}{0.4476} & \perc{0.7203}{0.9033} & \perc{0.7203}{0.9416} \\
Llama 3.3 70B & 0.7361 & \perc{0.7361}{0.7373} & \perc{0.7361}{0.4556} & \perc{0.7361}{0.8974} & \perc{0.7361}{0.9347} \\
Gemma 3 27b it & 0.7655 & \perc{0.7655}{0.7866} & \perc{0.7655}{0.6791} & \perc{0.7655}{0.8806} & \perc{0.7655}{0.9006} \\
Gemini 2.0 Flash & 0.7903 & \perc{0.7903}{0.7959} & \perc{0.7903}{0.6949} & \perc{0.7903}{0.8499} & \perc{0.7903}{0.8593} \\
\hline
\end{tabular}
\caption{F\textsubscript{1} scores for persuasion detection on German data sample of Persuaficial. More specifically, on sample of Persuaficial generated using German texts from SemEval 2023 Task 3 dataset. The first column reports performance on SemEval 2023 Task 3 German human-annotated texts. The remaining columns show performance on LLM-generated German counterparts.}
\label{tab:semeval_german_results}
\end{table*}

\begin{table*}[!ht]
\centering
\scriptsize
\begin{tabular}{lccccc}
\hline
\textbf{Classifier Models} & 
\textbf{Human-written} & \textbf{Paraphrasing Generation} & 
\multicolumn{2}{c}{\textbf{Rewriting Generation}} & 
\textbf{Open-ended Generation} \\ 
 & & & \textbf{Subtle Persuasion} & \textbf{Intensive Persuasion} & \\ \hline

\multicolumn{6}{l}{\textit{\textbf{Generating model: GPT 4.1 Mini}}} \\
GPT 4.1 Mini & 0.7505 & \perc{0.7505}{0.7356} & \perc{0.7505}{0.4568} & \perc{0.7505}{0.8021} & \perc{0.7505}{0.9251} \\
Llama 3.3 70B & 0.7605 & \perc{0.7605}{0.7460} & \perc{0.7605}{0.4675} & \perc{0.7605}{0.7996} & \perc{0.7605}{0.9164} \\
Gemma 3 27b it & 0.7827 & \perc{0.7827}{0.7733} & \perc{0.7827}{0.6489} & \perc{0.7827}{0.8190} & \perc{0.7827}{0.8816} \\
Gemini 2.0 Flash & 0.7812 & \perc{0.7812}{0.7812} & \perc{0.7812}{0.6705} & \perc{0.7812}{0.8189} & \perc{0.7812}{0.8418} \\
\hline
\multicolumn{6}{l}{\textit{\textbf{Generating model: Llama 3.3 70B}}} \\
GPT 4.1 Mini & 0.7505 & \perc{0.7505}{0.7160} & \perc{0.7505}{0.4548} & \perc{0.7505}{0.7959} & \perc{0.7505}{0.9251} \\
Llama 3.3 70B & 0.7605 & \perc{0.7605}{0.7216} & \perc{0.7605}{0.4794} & \perc{0.7605}{0.7996} & \perc{0.7605}{0.9174} \\
Gemma 3 27b it & 0.7827 & \perc{0.7827}{0.7744} & \perc{0.7827}{0.6908} & \perc{0.7827}{0.8343} & \perc{0.7827}{0.8826} \\
Gemini 2.0 Flash & 0.7812 & \perc{0.7812}{0.7758} & \perc{0.7812}{0.7116} & \perc{0.7812}{0.8269} & \perc{0.7812}{0.8418} \\
\hline
\multicolumn{6}{l}{\textit{\textbf{Generating model: Gemma 3 27b it}}} \\
GPT 4.1 Mini & 0.7505 & \perc{0.7505}{0.7492} & \perc{0.7505}{0.3376} & \perc{0.7505}{0.8848} & \perc{0.7505}{0.9251} \\
Llama 3.3 70B & 0.7605 & \perc{0.7605}{0.7366} & \perc{0.7605}{0.3520} & \perc{0.7605}{0.8815} & \perc{0.7605}{0.9174} \\
Gemma 3 27b it & 0.7827 & \perc{0.7827}{0.8068} & \perc{0.7827}{0.6179} & \perc{0.7827}{0.8676} & \perc{0.7827}{0.8826} \\
Gemini 2.0 Flash & 0.7812 & \perc{0.7812}{0.7898} & \perc{0.7812}{0.6523} & \perc{0.7812}{0.8388} & \perc{0.7812}{0.8418} \\
\hline
\multicolumn{6}{l}{\textit{\textbf{Generating model: Gemini 2.0 Flash}}} \\
GPT 4.1 Mini & 0.7505 & \perc{0.7505}{0.7807} & \perc{0.7505}{0.4670} & \perc{0.7505}{0.8996} & \perc{0.7505}{0.9251} \\
Llama 3.3 70B & 0.7605 & \perc{0.7605}{0.7759} & \perc{0.7605}{0.5120} & \perc{0.7605}{0.8920} & \perc{0.7605}{0.9174} \\
Gemma 3 27b it & 0.7827 & \perc{0.7827}{0.7920} & \perc{0.7827}{0.6771} & \perc{0.7827}{0.8696} & \perc{0.7827}{0.8826} \\
Gemini 2.0 Flash & 0.7812 & \perc{0.7812}{0.7972} & \perc{0.7812}{0.6858} & \perc{0.7812}{0.8408} & \perc{0.7812}{0.8418} \\
\hline
\end{tabular}
\caption{F\textsubscript{1} scores for persuasion detection on French data sample of Persuaficial. More specifically, on sample of Persuaficial generated using French texts from SemEval 2023 Task 3 dataset. The first column reports performance on SemEval 2023 Task 3 French human-annotated texts. The remaining columns show performance on LLM-generated French counterparts.}
\label{tab:semeval_french_results}
\end{table*}

\begin{table*}[!ht]
\centering
\scriptsize
\begin{tabular}{lccccc}
\hline
\textbf{Classifier Models} & 
\textbf{Human-written} & \textbf{Paraphrasing Generation} & 
\multicolumn{2}{c}{\textbf{Rewriting Generation}} & 
\textbf{Open-ended Generation} \\ 
 & & & \textbf{Subtle Persuasion} & \textbf{Intensive Persuasion} & \\ \hline

\multicolumn{6}{l}{\textit{\textbf{Generating model: GPT 4.1 Mini}}} \\
GPT 4.1 Mini & 0.7471 & \perc{0.7471}{0.7511} & \perc{0.7471}{0.4368} & \perc{0.7471}{0.7959} & \perc{0.7471}{0.9200} \\
Llama 3.3 70B & 0.7584 & \perc{0.7584}{0.7277} & \perc{0.7584}{0.4447} & \perc{0.7584}{0.7926} & \perc{0.7584}{0.9166} \\
Gemma 3 27b it & 0.7659 & \perc{0.7659}{0.7775} & \perc{0.7659}{0.6563} & \perc{0.7659}{0.8185} & \perc{0.7659}{0.8688} \\
Gemini 2.0 Flash & 0.7986 & \perc{0.7986}{0.7912} & \perc{0.7986}{0.6750} & \perc{0.7986}{0.8192} & \perc{0.7986}{0.8410} \\
\hline
\multicolumn{6}{l}{\textit{\textbf{Generating model: Llama 3.3 70B}}} \\
GPT 4.1 Mini & 0.7471 & \perc{0.7471}{0.7113} & \perc{0.7471}{0.4812} & \perc{0.7471}{0.7822} & \perc{0.7471}{0.9200} \\
Llama 3.3 70B & 0.7584 & \perc{0.7584}{0.6823} & \perc{0.7584}{0.4689} & \perc{0.7584}{0.7827} & \perc{0.7584}{0.9166} \\
Gemma 3 27b it & 0.7659 & \perc{0.7659}{0.7576} & \perc{0.7659}{0.6967} & \perc{0.7659}{0.8164} & \perc{0.7659}{0.8688} \\
Gemini 2.0 Flash & 0.7986 & \perc{0.7986}{0.7827} & \perc{0.7986}{0.7193} & \perc{0.7986}{0.8192} & \perc{0.7986}{0.8410} \\
\hline
\multicolumn{6}{l}{\textit{\textbf{Generating model: Gemma 3 27b it}}} \\
GPT 4.1 Mini & 0.7471 & \perc{0.7471}{0.7239} & \perc{0.7471}{0.3252} & \perc{0.7471}{0.8925} & \perc{0.7471}{0.9190} \\
Llama 3.3 70B & 0.7584 & \perc{0.7584}{0.7110} & \perc{0.7584}{0.3376} & \perc{0.7584}{0.8912} & \perc{0.7584}{0.9156} \\
Gemma 3 27b it & 0.7659 & \perc{0.7659}{0.7944} & \perc{0.7659}{0.6280} & \perc{0.7659}{0.8619} & \perc{0.7659}{0.8678} \\
Gemini 2.0 Flash & 0.7986 & \perc{0.7986}{0.7944} & \perc{0.7986}{0.6356} & \perc{0.7986}{0.8410} & \perc{0.7986}{0.8401} \\
\hline
\multicolumn{6}{l}{\textit{\textbf{Generating model: Gemini 2.0 Flash}}} \\
GPT 4.1 Mini & 0.7471 & \perc{0.7471}{0.7457} & \perc{0.7471}{0.4553} & \perc{0.7471}{0.9007} & \perc{0.7471}{0.9190} \\
Llama 3.3 70B & 0.7584 & \perc{0.7584}{0.7479} & \perc{0.7584}{0.4629} & \perc{0.7584}{0.9015} & \perc{0.7584}{0.9156} \\
Gemma 3 27b it & 0.7659 & \perc{0.7659}{0.7922} & \perc{0.7659}{0.6632} & \perc{0.7659}{0.8629} & \perc{0.7659}{0.8688} \\
Gemini 2.0 Flash & 0.7986 & \perc{0.7986}{0.8069} & \perc{0.7986}{0.6826} & \perc{0.7986}{0.8410} & \perc{0.7986}{0.8410} \\
\hline
\end{tabular}
\caption{F\textsubscript{1} scores for persuasion detection on Italian data sample of Persuaficial. More specifically, on sample of Persuaficial generated using Italian texts from SemEval 2023 Task 3 dataset. The first column reports performance on SemEval 2023 Task 3 Italian human-annotated texts. The remaining columns show performance on LLM-generated Italian counterparts.}
\label{tab:semeval_italian_results}
\end{table*}

\begin{table*}[!ht]
\centering
\scriptsize
\begin{tabular}{lccccc}
\hline
\textbf{Classifier Models} & 
\textbf{Human-written} & \textbf{Paraphrasing Generation} & 
\multicolumn{2}{c}{\textbf{Rewriting Generation}} & 
\textbf{Open-ended Generation} \\ 
 & & & \textbf{Subtle Persuasion} & \textbf{Intensive Persuasion} & \\ \hline

\multicolumn{6}{l}{\textit{\textbf{Generating model: GPT 4.1 Mini}}} \\
GPT 4.1 Mini & 0.7330 & \perc{0.7330}{0.7010} & \perc{0.7330}{0.4634} & \perc{0.7330}{0.7936} & \perc{0.7330}{0.9372} \\
Llama 3.3 70B & 0.7676 & \perc{0.7676}{0.7398} & \perc{0.7676}{0.5165} & \perc{0.7676}{0.8016} & \perc{0.7676}{0.9208} \\
Gemma 3 27b it & 0.7728 & \perc{0.7728}{0.7752} & \perc{0.7728}{0.6942} & \perc{0.7728}{0.8075} & \perc{0.7728}{0.8834} \\
Gemini 2.0 Flash & 0.7733 & \perc{0.7733}{0.7744} & \perc{0.7733}{0.7151} & \perc{0.7733}{0.8020} & \perc{0.7733}{0.8217} \\
\hline
\multicolumn{6}{l}{\textit{\textbf{Generating model: Llama 3.3 70B}}} \\
GPT 4.1 Mini & 0.7330 & \perc{0.7330}{0.6829} & \perc{0.7330}{0.5215} & \perc{0.7330}{0.7793} & \perc{0.7330}{0.9352} \\
Llama 3.3 70B & 0.7676 & \perc{0.7676}{0.7177} & \perc{0.7676}{0.5637} & \perc{0.7676}{0.7880} & \perc{0.7676}{0.9198} \\
Gemma 3 27b it & 0.7728 & \perc{0.7728}{0.7596} & \perc{0.7728}{0.7232} & \perc{0.7728}{0.8176} & \perc{0.7728}{0.8834} \\
Gemini 2.0 Flash & 0.7733 & \perc{0.7733}{0.7563} & \perc{0.7733}{0.7321} & \perc{0.7733}{0.7980} & \perc{0.7733}{0.8217} \\
\hline
\multicolumn{6}{l}{\textit{\textbf{Generating model: Gemma 3 27b it}}} \\
GPT 4.1 Mini & 0.7330 & \perc{0.7330}{0.7143} & \perc{0.7330}{0.3777} & \perc{0.7330}{0.9086} & \perc{0.7330}{0.9372} \\
Llama 3.3 70B & 0.7676 & \perc{0.7676}{0.7330} & \perc{0.7676}{0.4311} & \perc{0.7676}{0.9047} & \perc{0.7676}{0.9208} \\
Gemma 3 27b it & 0.7728 & \perc{0.7728}{0.7869} & \perc{0.7728}{0.6737} & \perc{0.7728}{0.8754} & \perc{0.7728}{0.8834} \\
Gemini 2.0 Flash & 0.7733 & \perc{0.7733}{0.7765} & \perc{0.7733}{0.6623} & \perc{0.7733}{0.8198} & \perc{0.7733}{0.8217} \\
\hline
\multicolumn{6}{l}{\textit{\textbf{Generating model: Gemini 2.0 Flash}}} \\
GPT 4.1 Mini & 0.7330 & \perc{0.7330}{0.7258} & \perc{0.7330}{0.4696} & \perc{0.7330}{0.9117} & \perc{0.7330}{0.9372} \\
Llama 3.3 70B & 0.7676 & \perc{0.7676}{0.7650} & \perc{0.7676}{0.5051} & \perc{0.7676}{0.9128} & \perc{0.7676}{0.9208} \\
Gemma 3 27b it & 0.7728 & \perc{0.7728}{0.7916} & \perc{0.7728}{0.6764} & \perc{0.7728}{0.8704} & \perc{0.7728}{0.8834} \\
Gemini 2.0 Flash & 0.7733 & \perc{0.7733}{0.7858} & \perc{0.7733}{0.6975} & \perc{0.7733}{0.8207} & \perc{0.7733}{0.8217} \\
\hline
\end{tabular}
\caption{F\textsubscript{1} scores for persuasion detection on Polish data sample of Persuaficial. More specifically, on sample of Persuaficial generated using Polish texts from SemEval 2023 Task 3 dataset. The first column reports performance on SemEval 2023 Task 3 Polish human-annotated texts. The remaining columns show performance on LLM-generated Polish counterparts.}
\label{tab:semeval_polish_results}
\end{table*}

\begin{table*}[!ht]
\centering
\scriptsize
\begin{tabular}{lccccc}
\hline
\textbf{Classifier Models} & 
\textbf{Human-written} & \textbf{Paraphrasing Generation} & 
\multicolumn{2}{c}{\textbf{Rewriting Generation}} & 
\textbf{Open-ended Generation} \\ 
 & & & \textbf{Subtle Persuasion} & \textbf{Intensive Persuasion} & \\ \hline

\multicolumn{6}{l}{\textit{\textbf{Generating model: GPT 4.1 Mini}}} \\
GPT 4.1 Mini & 0.7246 & \perc{0.7246}{0.6889} & \perc{0.7246}{0.4464} & \perc{0.7246}{0.7844} & \perc{0.7246}{0.9017} \\
Llama 3.3 70B & 0.7408 & \perc{0.7408}{0.7166} & \perc{0.7408}{0.4556} & \perc{0.7408}{0.7915} & \perc{0.7408}{0.9071} \\
Gemma 3 27b it & 0.7360 & \perc{0.7360}{0.7396} & \perc{0.7360}{0.6069} & \perc{0.7360}{0.7843} & \perc{0.7360}{0.8562} \\
Gemini 2.0 Flash & 0.7683 & \perc{0.7683}{0.7571} & \perc{0.7683}{0.6767} & \perc{0.7683}{0.7784} & \perc{0.7683}{0.8019} \\
\hline
\multicolumn{6}{l}{\textit{\textbf{Generating model: Llama 3.3 70B}}} \\
GPT 4.1 Mini & 0.7246 & \perc{0.7246}{0.6889} & \perc{0.7246}{0.4794} & \perc{0.7246}{0.7546} & \perc{0.7246}{0.9017} \\
Llama 3.3 70B & 0.7408 & \perc{0.7408}{0.6740} & \perc{0.7408}{0.4635} & \perc{0.7408}{0.7603} & \perc{0.7408}{0.9091} \\
Gemma 3 27b it & 0.7360 & \perc{0.7360}{0.7203} & \perc{0.7360}{0.6311} & \perc{0.7360}{0.7732} & \perc{0.7360}{0.8562} \\
Gemini 2.0 Flash & 0.7683 & \perc{0.7683}{0.7477} & \perc{0.7683}{0.6986} & \perc{0.7683}{0.7774} & \perc{0.7683}{0.8019} \\
\hline
\multicolumn{6}{l}{\textit{\textbf{Generating model: Gemma 3 27b it}}} \\
GPT 4.1 Mini & 0.7246 & \perc{0.7246}{0.7138} & \perc{0.7246}{0.3651} & \perc{0.7246}{0.8805} & \perc{0.7246}{0.9017} \\
Llama 3.3 70B & 0.7408 & \perc{0.7408}{0.7248} & \perc{0.7408}{0.3562} & \perc{0.7408}{0.8858} & \perc{0.7408}{0.9091} \\
Gemma 3 27b it & 0.7360 & \perc{0.7360}{0.7491} & \perc{0.7360}{0.5907} & \perc{0.7360}{0.8453} & \perc{0.7360}{0.8562} \\
Gemini 2.0 Flash & 0.7683 & \perc{0.7683}{0.7704} & \perc{0.7683}{0.6577} & \perc{0.7683}{0.7971} & \perc{0.7683}{0.8019} \\
\hline
\multicolumn{6}{l}{\textit{\textbf{Generating model: Gemini 2.0 Flash}}} \\
GPT 4.1 Mini & 0.7246 & \perc{0.7246}{0.7378} & \perc{0.7246}{0.4660} & \perc{0.7246}{0.8774} & \perc{0.7246}{0.9017} \\
Llama 3.3 70B & 0.7408 & \perc{0.7408}{0.7500} & \perc{0.7408}{0.4496} & \perc{0.7408}{0.8920} & \perc{0.7408}{0.9091} \\
Gemma 3 27b it & 0.7360 & \perc{0.7360}{0.7572} & \perc{0.7360}{0.6227} & \perc{0.7360}{0.8362} & \perc{0.7360}{0.8562} \\
Gemini 2.0 Flash & 0.7683 & \perc{0.7683}{0.7714} & \perc{0.7683}{0.6849} & \perc{0.7683}{0.7981} & \perc{0.7683}{0.8019} \\
\hline
\end{tabular}
\caption{F\textsubscript{1} scores for persuasion detection on Russian data sample of Persuaficial. More specifically, on sample of Persuaficial generated using Russian texts from SemEval 2023 Task 3 dataset. The first column reports performance on SemEval 2023 Task 3 Russian human-annotated texts. The remaining columns show performance on LLM-generated Russian counterparts.}
\label{tab:semeval_russian_results}
\end{table*}

\section{StyloMetrix and Its Usefulness in AI vs. Human Persuasive Text Analysis}
\label{app:stylometrix_justification}

To further demonstrate the utility of StyloMetrix for analyzing human-written versus AI-generated persuasive texts, we conduct a classification study using classical machine learning models with features calculated by StyloMetrix. We aimed to prove that linguistic features contain enough information to differentiate AI-generated persuasion from human-written persuasion.

For all GPT 4.1. Mini-generated English synthetic texts produced with each generation approach, we trained a separate classifier.
For each experiemnt, we split the data into training and test sets, allocating 70\% for training and 30\% for testing. To ensure a credible evaluation, each human-written text and its LLM-generated counterpart were placed in the same split, either training or test. This prevents the classifier from exploiting the potential direct similarities between the paired texts. We employed widely used tree-based machine learning methods as classifiers, as they naturally capture non-linear interactions and are well-suited for moderate- to high-dimensional tabular data \cite{grinsztajn2022tree, uddin2024confirming}. Previous work has shown that, for tabular data, tree-based models can even outperform deep learning approaches \cite{grinsztajn2022tree}. 

Table \ref{tab:tree_performance} shows the results of these experiments. The outcomes show a clear progression: the more generative freedom the LLM is given, the easier it becomes for tree-ensemble models to distinguish its outputs from human-written persuasion. \textit{Paraphrasing} keeps the AI text close to the original human style, yielding only moderate detection performance. In the \textit{Rewriting} conditions, the model introduces larger stylistic shifts—whether by making persuasion subtler or more intense—which improves separability. \textit{Open-ended} generation, starting from only a brief neutral summary, produces the greatest stylistic divergence and thus the highest classification accuracy. Overall, stylistic features become increasingly informative as the generation task becomes less constrained.

\begin{table*}[!ht]
\centering
\scriptsize
\begin{tabular}{lcccc}
\hline
\textbf{Model} & \textbf{Precision} & \textbf{Recall} & \textbf{F1} & \textbf{Accuracy} \\
\hline

\multicolumn{5}{l}{\textit{\textbf{GPT 4.1 Mini -- Paraphrasing}}} \\
RF   & 0.74 & 0.74 & 0.74 & 0.74 \\
XGB  & 0.76 & 0.76 & 0.76 & 0.76 \\
LGBM & 0.76 & 0.76 & 0.76 & 0.76 \\
\hline

\multicolumn{5}{l}{\textit{\textbf{GPT 4.1 Mini -- Rewriting with Subtle Persuasive Effect}}} \\
RF   & 0.84 & 0.84 & 0.84 & 0.84 \\
XGB  & 0.87 & 0.87 & 0.87 & 0.87 \\
LGBM & 0.86 & 0.86 & 0.86 & 0.86 \\
\hline

\multicolumn{5}{l}{\textit{\textbf{GPT 4.1 Mini -- Rewriting with Intensified Persuasive Effect}}} \\
RF   & 0.83 & 0.83 & 0.83 & 0.83 \\
XGB  & 0.84 & 0.84 & 0.84 & 0.84 \\
LGBM & 0.86 & 0.85 & 0.85 & 0.85 \\
\hline

\multicolumn{5}{l}{\textit{\textbf{GPT 4.1 Mini -- Open-ended}}} \\
RF   & 0.97 & 0.97 & 0.97 & 0.97 \\
XGB  & 0.97 & 0.97 & 0.97 & 0.97 \\
LGBM & 0.98 & 0.98 & 0.98 & 0.98 \\
\hline

\end{tabular}
\caption{Classification performance of detecting GPT-4.1-Mini-generated persuasive texts versus human-written persuasive texts using linguistic-feature representations. Each experiment reflects a different AI-generation strategy and uses data combined from three English datasets.}
\label{tab:tree_performance}
\end{table*}

\section{Statistical Analysis with Wilcoxon Signed-rank Test}
\label{app:wilcoxon_signed-rank}

To assess whether the observed shifts are statistically significant without assuming normality of the paired differences, we apply the Wilcoxon signed-rank test:

\begin{equation}
H_0: \tilde{\mu}_{\Delta} = 0 \quad \text{vs.} \quad H_1: \tilde{\mu}_{\Delta} \neq 0
\end{equation}
where $\tilde{\mu}_{\Delta} = \text{median}(\Delta_j^{1}, \Delta_j^{2}, \ldots, \Delta_j^{n})$ is the population median of the paired differences for feature $j$. The test operates by ranking the absolute values of the non-zero paired differences, then comparing the sum of ranks associated with positive differences against those associated with negative differences. This non-parametric approach is robust to outliers and skewed distributions, making it well-suited for stylometric features that may not follow a Gaussian distribution.  As we perform one test per feature, we control for multiple comparisons
using the Benjamini-Hochberg false discovery rate (FDR) correction \cite{benjamini1995controlling} across all features. We report a significance indicator in our tables.

Finally, features are ranked by the absolute value of their paired Cohen's $d$, and the top-ranked features are reported along with their FDR-corrected significance status. This ranking highlights the stylometric dimensions along which AI-generated persuasive text diverges most strongly from its human-written counterpart.




\section{Linguistic Differences between Human and Machine Generated Persuasive Texts - Additional Results}
\label{app:ling_diff_for_add_results}

Tables \ref{tab:stylometrix_paraphrased_gpt-4.1-mini_top20} - \ref{tab:stylometrix_open_end_gemini_top20} report the top features that most strongly distinguish AI-generated persuasive texts from human-written persuasive texts. We provide 16 tables in total, reflecting Cohen’s d effect sizes computed separately for each generating model and for each generation setting: Paraphrasing, Rewriting subtle persuasion, Rewriting intensified persuasion, and Open-ended generation.

\begin{table}[!ht]
\centering
\scriptsize
\begin{tabular}{lcc}
\hline
\textbf{Stylometric Feature} & \textbf{Cohen's d} & \textbf{Sig.} \\
\hline
\multicolumn{3}{l}{\textit{\textbf{GPT-4.1-mini - Paraphrased}}} \\
L\_TYPE\_TOKEN\_RATIO\_LEMMAS & 0.9218 & \checkmark \\
L\_CONT\_T & 0.8239 & \checkmark \\
ST\_REPETITIONS\_WORDS & -0.7487 & \checkmark \\
L\_CONT\_A & 0.6522 & \checkmark \\
L\_FUNC\_A & -0.5824 & \checkmark \\
SY\_INVERSE\_PATTERNS & -0.4217 & \checkmark \\
L\_PUNCT\_COM & 0.3217 & \checkmark \\
L\_PUNCT\_DASH & 0.3116 & \checkmark \\
SENT\_D\_NP & 0.3055 & \checkmark \\
L\_LINKS & -0.2848 & \checkmark \\
L\_PLURAL\_NOUNS & 0.2717 & \checkmark \\
L\_ADJ\_POSITIVE & 0.2461 & \checkmark \\
SENT\_ST\_DIFFERENCE & -0.2374 & \checkmark \\
POS\_PRO & -0.2357 & \checkmark \\
L\_YOU\_PRON & -0.2195 & \checkmark \\
L\_SECOND\_PERSON\_PRON & -0.2118 & \checkmark \\
POS\_ADJ & 0.2080 & \checkmark \\
L\_PUNCT\_SEMC & -0.1991 & \checkmark \\
ASM & -0.1920 & \checkmark \\
L\_ADV\_SUPERLATIVE & 0.1919 & \checkmark \\
\hline
\end{tabular}
\caption{Top 20 linguistic features for AI-generated persuasive text with \textit{Paraphrasing} generation approach and GPT 4.1 Mini model vs. human-written persuasive texts (three samples of human datasets combined vs. AI counterparts). Cohen's d sorted by absolute value.}
\label{tab:stylometrix_paraphrased_gpt-4.1-mini_top20}
\end{table}

\begin{table}[!ht]
\centering
\scriptsize
\begin{tabular}{lcc}
\hline
\textbf{Stylometric Feature} & \textbf{Cohen's d} & \textbf{Sig.} \\
\hline
\multicolumn{3}{l}{\textit{\textbf{GPT 4.1 Mini - Rewriting with Subtle Persuasive Effect}}} \\
L\_TYPE\_TOKEN\_RATIO\_LEMMAS & 0.9407 & \checkmark \\
L\_CONT\_T & 0.8198 & \checkmark \\
ST\_REPETITIONS\_WORDS & -0.7602 & \checkmark \\
L\_PLURAL\_NOUNS & 0.6603 & \checkmark \\
L\_CONT\_A & 0.6331 & \checkmark \\
L\_FUNC\_A & -0.6062 & \checkmark \\
POS\_PRO & -0.5107 & \checkmark \\
SY\_INVERSE\_PATTERNS & -0.4739 & \checkmark \\
SY\_NARRATIVE & 0.4704 & \checkmark \\
L\_SECOND\_PERSON\_PRON & -0.4526 & \checkmark \\
VT\_MIGHT & 0.4523 & \checkmark \\
SENT\_D\_PP & 0.4265 & \checkmark \\
G\_ACTIVE & -0.4122 & \checkmark \\
SENT\_ST\_DIFFERENCE & -0.4109 & \checkmark \\
L\_YOU\_PRON & -0.3956 & \checkmark \\
L\_I\_PRON & -0.3724 & \checkmark \\
L\_FIRST\_PERSON\_SING\_PRON & -0.3724 & \checkmark \\
CDS & -0.3685 & \checkmark \\
POS\_NOUN & 0.3636 & \checkmark \\
G\_FUTURE & -0.3284 & \checkmark \\
\hline
\end{tabular}
\caption{Top 20 linguistic features for AI-generated persuasive text with \textit{Rewriting with Subtle Persuasive Effect} generation approach and GPT 4.1 Mini model vs. human-written persuasive texts (three samples of human datasets combined vs. AI counterparts). Cohen's d sorted by absolute value.}
\label{tab:stylometrix_rewritten_gpt-4.1-mini_top20}
\end{table}

\begin{table}[!ht]
\centering
\scriptsize
\begin{tabular}{lcc}
\hline
\textbf{Stylometric Feature} & \textbf{Cohen's d} & \textbf{Sig.} \\
\hline
\multicolumn{3}{l}{\textit{\textbf{GPT 4.1 Mini - Rewriting with Intensified Persuasive Effect}}} \\
L\_CONT\_T & 1.1821 & \checkmark \\
L\_TYPE\_TOKEN\_RATIO\_LEMMAS & 1.0916 & \checkmark \\
L\_CONT\_A & 1.0240 & \checkmark \\
L\_FUNC\_A & -0.8593 & \checkmark \\
ST\_REPETITIONS\_WORDS & -0.8215 & \checkmark \\
L\_PUNCT\_DASH & 0.5816 & \checkmark \\
L\_ADJ\_POSITIVE & 0.5443 & \checkmark \\
POS\_ADJ & 0.5200 & \checkmark \\
L\_ADV\_SUPERLATIVE & 0.4963 & \checkmark \\
POS\_ADV & 0.4924 & \checkmark \\
L\_ADV\_COMPARATIVE & 0.4869 & \checkmark \\
SY\_INVERSE\_PATTERNS & -0.4506 & \checkmark \\
POS\_PRO & -0.4148 & \checkmark \\
L\_ADV\_POSITIVE & 0.3644 & \checkmark \\
L\_PUNCT\_COM & 0.3578 & \checkmark \\
SENT\_D\_NP & 0.3522 & \checkmark \\
L\_NOUN\_PHRASES & 0.3414 & \checkmark \\
PS\_CAUSE & -0.3385 & \checkmark \\
L\_LINKS & -0.3305 & \checkmark \\
SENT\_ST\_WRDSPERSENT & 0.3042 & \checkmark \\
\hline
\end{tabular}
\caption{Top 20 linguistic features for AI-generated persuasive text with \textit{Rewriting with Intensified Persuasive Effect} generation approach and GPT 4.1 Mini model vs. human-written persuasive texts (three samples of human datasets combined vs. AI counterparts). Cohen's d sorted by absolute value.}
\label{tab:stylometrix_intensified_gpt-4.1-mini_top20}
\end{table}

\begin{table}[!ht]
\centering
\scriptsize
\begin{tabular}{lcc}
\hline
\textbf{Stylometric Feature} & \textbf{Cohen's d} & \textbf{Sig.} \\
\hline
\multicolumn{3}{l}{\textit{\textbf{GPT 4.1 Mini - Open-ended}}} \\
L\_CONT\_T & 1.4059 & \checkmark \\
L\_CONT\_A & 1.2264 & \checkmark \\
L\_TYPE\_TOKEN\_RATIO\_LEMMAS & 1.0753 & \checkmark \\
ST\_REPETITIONS\_WORDS & -0.8541 & \checkmark \\
SENT\_D\_NP & 0.7600 & \checkmark \\
L\_FUNC\_A & -0.7259 & \checkmark \\
POS\_NOUN & 0.7240 & \checkmark \\
L\_PUNCT\_DASH & 0.7067 & \checkmark \\
POS\_ADJ & 0.6430 & \checkmark \\
L\_ADJ\_POSITIVE & 0.6377 & \checkmark \\
SY\_IMPERATIVE & 0.5960 & \checkmark \\
VF\_INFINITIVE & 0.5668 & \checkmark \\
G\_PAST & -0.5589 & \checkmark \\
SY\_INVERSE\_PATTERNS & -0.5410 & \checkmark \\
L\_LINKS & -0.5285 & \checkmark \\
G\_ACTIVE & -0.5023 & \checkmark \\
SENT\_D\_VP & 0.4973 & \checkmark \\
VT\_PAST\_SIMPLE & -0.4660 & \checkmark \\
L\_SINGULAR\_NOUNS & 0.4653 & \checkmark \\
L\_I\_PRON & -0.4632 & \checkmark \\
\hline
\end{tabular}
\caption{Top 20 linguistic features for AI-generated persuasive text with \textit{Open-ended} generation approach and GPT 4.1 Mini model vs. human-written persuasive texts (three samples of human datasets combined vs. AI counterparts). Cohen's d sorted by absolute value.}
\label{tab:stylometrix_open_end_gpt-4.1-mini_top20}
\end{table}

\begin{table}[!ht]
\centering
\scriptsize
\begin{tabular}{lcc}
\hline
\textbf{Stylometric Feature} & \textbf{Cohen's d} & \textbf{Sig.} \\
\hline
\multicolumn{3}{l}{\textit{\textbf{Llama - Paraphrasing}}} \\
SENT\_ST\_WRDSPERSENT & 0.7721 & \checkmark \\
SENT\_ST\_DIFFERENCE & -0.7695 & \checkmark \\
L\_CONT\_T & 0.7592 & \checkmark \\
L\_PUNCT\_COM & 0.6832 & \checkmark \\
L\_TYPE\_TOKEN\_RATIO\_LEMMAS & 0.6077 & \checkmark \\
L\_CONT\_A & 0.5676 & \checkmark \\
L\_PUNCT\_DOT & -0.5399 & \checkmark \\
POS\_PRO & -0.5077 & \checkmark \\
SY\_INVERSE\_PATTERNS & -0.4720 & \checkmark \\
G\_ACTIVE & -0.4647 & \checkmark \\
L\_FUNC\_A & -0.4470 & \checkmark \\
L\_SECOND\_PERSON\_PRON & -0.4128 & \checkmark \\
L\_YOU\_PRON & -0.4097 & \checkmark \\
ST\_REPETITIONS\_WORDS & -0.3859 & \checkmark \\
L\_ADJ\_POSITIVE & 0.3841 & \checkmark \\
VT\_PRESENT\_SIMPLE & -0.3681 & \checkmark \\
L\_PROPER\_NAME & -0.3615 & \checkmark \\
L\_LINKS & -0.3494 & \checkmark \\
POS\_NOUN & 0.3463 & \checkmark \\
ASM & -0.3324 & \checkmark \\
\hline
\end{tabular}
\caption{Top 20 linguistic features for AI-generated persuasive text with \textit{Paraphrasing} generation approach and Llama 3.3 70B model vs. human-written persuasive texts (three samples of human datasets combined vs. AI counterparts). Cohen's d sorted by absolute value.}
\label{tab:stylometrix_paraphrased_llama_top20}
\end{table}

\begin{table}[!ht]
\centering
\scriptsize
\begin{tabular}{lcc}
\hline
\textbf{Stylometric Feature} & \textbf{Cohen's d} & \textbf{Sig.} \\
\hline
\multicolumn{3}{l}{\textit{\textbf{Llama - Rewriting with Subtle Persuasive Effect}}} \\
SENT\_ST\_WRDSPERSENT & 0.7274 & \checkmark \\
L\_CONT\_T & 0.7048 & \checkmark \\
SENT\_D\_PP & 0.6735 & \checkmark \\
SENT\_ST\_DIFFERENCE & -0.6391 & \checkmark \\
L\_PROPER\_NAME & -0.6114 & \checkmark \\
POS\_NOUN & 0.6002 & \checkmark \\
G\_ACTIVE & -0.5923 & \checkmark \\
SY\_INVERSE\_PATTERNS & -0.5443 & \checkmark \\
L\_ADJ\_POSITIVE & 0.5337 & \checkmark \\
L\_PUNCT\_COM & 0.5158 & \checkmark \\
POS\_PRO & -0.5124 & \checkmark \\
L\_CONT\_A & 0.5080 & \checkmark \\
L\_SECOND\_PERSON\_PRON & -0.5064 & \checkmark \\
L\_PLURAL\_NOUNS & 0.5021 & \checkmark \\
POS\_PREP & 0.4996 & \checkmark \\
POS\_ADJ & 0.4885 & \checkmark \\
L\_TYPE\_TOKEN\_RATIO\_LEMMAS & 0.4744 & \checkmark \\
L\_YOU\_PRON & -0.4730 & \checkmark \\
SY\_NARRATIVE & 0.4673 & \checkmark \\
SY\_SUBORD\_SENT & 0.4254 & \checkmark \\
\hline
\end{tabular}
\caption{Top 20 linguistic features for AI-generated persuasive text with \textit{Rewriting with Subtle Persuasive Effect} generation approach and Llama 3.3 70B model vs. human-written persuasive texts (three samples of human datasets combined vs. AI counterparts). Cohen's d sorted by absolute value.}
\label{tab:stylometrix_rewritten_llama_top20}
\end{table}

\begin{table}[!ht]
\centering
\scriptsize
\begin{tabular}{lcc}
\hline
\textbf{Stylometric Feature} & \textbf{Cohen's d} & \textbf{Sig.} \\
\hline
\multicolumn{3}{l}{\textit{\textbf{Llama - Rewriting with Intensified Persuasive Effect}}} \\
L\_CONT\_T & 0.9518 & \checkmark \\
SENT\_ST\_WRDSPERSENT & 0.8794 & \checkmark \\
L\_CONT\_A & 0.7635 & \checkmark \\
L\_ADJ\_POSITIVE & 0.7466 & \checkmark \\
SENT\_ST\_DIFFERENCE & -0.7274 & \checkmark \\
L\_PUNCT\_COM & 0.7113 & \checkmark \\
POS\_ADJ & 0.6743 & \checkmark \\
L\_FUNC\_T & -0.6440 & \checkmark \\
G\_ACTIVE & -0.6295 & \checkmark \\
L\_FUNC\_A & -0.6168 & \checkmark \\
L\_PUNCT\_DOT & -0.5948 & \checkmark \\
L\_PROPER\_NAME & -0.5737 & \checkmark \\
L\_TYPE\_TOKEN\_RATIO\_LEMMAS & 0.4713 & \checkmark \\
POS\_PRO & -0.4696 & \checkmark \\
G\_PAST & -0.4403 & \checkmark \\
SY\_INVERSE\_PATTERNS & -0.4362 & \checkmark \\
L\_SECOND\_PERSON\_PRON & -0.4237 & \checkmark \\
L\_NOUN\_PHRASES & 0.4208 & \checkmark \\
L\_YOU\_PRON & -0.4169 & \checkmark \\
SENT\_D\_VP & 0.4091 & \checkmark \\
\hline
\end{tabular}
\caption{Top 20 linguistic features for AI-generated persuasive text with \textit{Rewriting with Intensified Persuasive Effect} generation approach and Llama 3.3 70B model vs. human-written persuasive texts (three samples of human datasets combined vs. AI counterparts). Cohen's d sorted by absolute value.}
\label{tab:stylometrix_intensified_llama_top20}
\end{table}

\begin{table}[!ht]
\centering
\scriptsize
\begin{tabular}{lcc}
\hline
\textbf{Stylometric Feature} & \textbf{Cohen's d} & \textbf{Sig.} \\
\hline
\multicolumn{3}{l}{\textit{\textbf{Llama - Open-ended}}} \\
VF\_INFINITIVE & 1.0143 & \checkmark \\
SY\_IMPERATIVE & 0.8892 & \checkmark \\
G\_ACTIVE & -0.6972 & \checkmark \\
SENT\_D\_VP & 0.6751 & \checkmark \\
L\_OUR\_PRON & 0.6048 & \checkmark \\
G\_PAST & -0.6043 & \checkmark \\
L\_CONT\_T & 0.5952 & \checkmark \\
L\_PUNCT\_DOT & -0.5939 & \checkmark \\
VT\_PAST\_SIMPLE & -0.5411 & \checkmark \\
L\_LINKS & -0.5285 & \checkmark \\
SY\_EXCLAMATION & 0.5258 & \checkmark \\
SY\_COORD\_SENT & 0.5123 & \checkmark \\
SY\_INVERSE\_PATTERNS & -0.5000 & \checkmark \\
L\_PUNCT & -0.4763 & \checkmark \\
SENT\_D\_NP & 0.4762 & \checkmark \\
L\_WE\_PRON & 0.4659 & \checkmark \\
L\_I\_PRON & -0.4463 & \checkmark \\
L\_FIRST\_PERSON\_SING\_PRON & -0.4463 & \checkmark \\
POS\_NOUN & 0.4267 & \checkmark \\
L\_PERSONAL\_NAME & -0.4258 & \checkmark \\
\hline
\end{tabular}
\caption{Top 20 linguistic features for AI-generated persuasive text with \textit{Open-ended} generation approach and Llama 3.3 70B model vs. human-written persuasive texts (three samples of human datasets combined vs. AI counterparts). Cohen's d sorted by absolute value.}
\label{tab:stylometrix_open_end_llama_top20}
\end{table}

\begin{table}[!ht]
\centering
\scriptsize
\begin{tabular}{lcc}
\hline
\textbf{Stylometric Feature} & \textbf{Cohen's d} & \textbf{Sig.} \\
\hline
\multicolumn{3}{l}{\textit{\textbf{Gemma - Paraphrasing}}} \\
L\_CONT\_T & 0.8744 & \checkmark \\
L\_TYPE\_TOKEN\_RATIO\_LEMMAS & 0.7274 & \checkmark \\
L\_FUNC\_A & -0.6963 & \checkmark \\
L\_CONT\_A & 0.6769 & \checkmark \\
ST\_REPETITIONS\_WORDS & -0.6665 & \checkmark \\
SY\_INVERSE\_PATTERNS & -0.4798 & \checkmark \\
L\_ADV\_SUPERLATIVE & 0.3887 & \checkmark \\
L\_ADV\_COMPARATIVE & 0.3845 & \checkmark \\
SENT\_ST\_WRDSPERSENT & 0.3600 & \checkmark \\
PS\_CONDITION & -0.3599 & \checkmark \\
POS\_PRO & -0.3556 & \checkmark \\
POS\_ADV & 0.3344 & \checkmark \\
L\_PUNCT\_COM & 0.3338 & \checkmark \\
CDS & -0.3291 & \checkmark \\
L\_ADJ\_POSITIVE & 0.3191 & \checkmark \\
L\_SECOND\_PERSON\_PRON & -0.3096 & \checkmark \\
SENT\_ST\_DIFFERENCE & -0.3074 & \checkmark \\
L\_FUNC\_T & -0.3063 & \checkmark \\
L\_YOU\_PRON & -0.2964 & \checkmark \\
ASM & -0.2902 & \checkmark \\
\hline
\end{tabular}
\caption{Top 20 linguistic features for AI-generated persuasive text with \textit{Paraphrasing} generation approach and Gemma 3 27b it model vs. human-written persuasive texts (three samples of human datasets combined vs. AI counterparts). Cohen's d sorted by absolute value.}
\label{tab:stylometrix_paraphrased_gemma_top20}
\end{table}

\begin{table}[!ht]
\centering
\scriptsize
\begin{tabular}{lcc}
\hline
\textbf{Stylometric Feature} & \textbf{Cohen's d} & \textbf{Sig.} \\
\hline
\multicolumn{3}{l}{\textit{\textbf{Gemma - Rewriting with Subtle Persuasive Effect}}} \\
L\_CONT\_T & 1.0167 & \checkmark \\
L\_CONT\_A & 0.8716 & \checkmark \\
L\_TYPE\_TOKEN\_RATIO\_LEMMAS & 0.8497 & \checkmark \\
L\_FUNC\_A & -0.7555 & \checkmark \\
ST\_REPETITIONS\_WORDS & -0.6725 & \checkmark \\
POS\_NOUN & 0.6419 & \checkmark \\
POS\_PRO & -0.5732 & \checkmark \\
SY\_INVERSE\_PATTERNS & -0.5399 & \checkmark \\
L\_PLURAL\_NOUNS & 0.5274 & \checkmark \\
G\_ACTIVE & -0.5179 & \checkmark \\
SENT\_D\_PP & 0.5060 & \checkmark \\
L\_SECOND\_PERSON\_PRON & -0.4606 & \checkmark \\
SENT\_D\_NP & 0.4539 & \checkmark \\
CDS & -0.4498 & \checkmark \\
L\_YOU\_PRON & -0.4263 & \checkmark \\
L\_ADJ\_POSITIVE & 0.4165 & \checkmark \\
L\_PUNCT\_COM & 0.3967 & \checkmark \\
L\_NOUN\_PHRASES & 0.3906 & \checkmark \\
SY\_NARRATIVE & 0.3875 & \checkmark \\
VT\_MIGHT & 0.3785 & \checkmark \\
\hline
\end{tabular}
\caption{Top 20 linguistic features for AI-generated persuasive text with \textit{Rewriting with Subtle Persuasive Effect} generation approach and Gemma 3 27b it model vs. human-written persuasive texts (three samples of human datasets combined vs. AI counterparts). Cohen's d sorted by absolute value.}
\label{tab:stylometrix_rewritten_gemma_top20}
\end{table}

\begin{table}[!ht]
\centering
\scriptsize
\begin{tabular}{lcc}
\hline
\textbf{Stylometric Feature} & \textbf{Cohen's d} & \textbf{Sig.} \\
\hline
\multicolumn{3}{l}{\textit{\textbf{Gemma - Rewriting with Intensified Persuasive Effect}}} \\
L\_CONT\_T & 1.1099 & \checkmark \\
L\_FUNC\_A & -0.9504 & \checkmark \\
L\_CONT\_A & 0.9499 & \checkmark \\
L\_ADJ\_POSITIVE & 0.8330 & \checkmark \\
POS\_ADJ & 0.7743 & \checkmark \\
L\_FUNC\_T & -0.7336 & \checkmark \\
L\_TYPE\_TOKEN\_RATIO\_LEMMAS & 0.5909 & \checkmark \\
ST\_REPETITIONS\_WORDS & -0.5650 & \checkmark \\
PS\_CONDITION & -0.5605 & \checkmark \\
SENT\_D\_NP & 0.5390 & \checkmark \\
G\_ACTIVE & -0.5361 & \checkmark \\
POS\_PRO & -0.5318 & \checkmark \\
L\_NOUN\_PHRASES & 0.5293 & \checkmark \\
PS\_CAUSE & -0.5251 & \checkmark \\
L\_PROPER\_NAME & -0.4946 & \checkmark \\
CDS & -0.4634 & \checkmark \\
L\_SINGULAR\_NOUNS & 0.4587 & \checkmark \\
POS\_NOUN & 0.4537 & \checkmark \\
SY\_INVERSE\_PATTERNS & -0.4488 & \checkmark \\
L\_ADV\_SUPERLATIVE & 0.4420 & \checkmark \\
\hline
\end{tabular}
\caption{Top 20 linguistic features for AI-generated persuasive text with \textit{Rewriting with Intensified Persuasive Effect} generation approach and Gemma 3 27b it model vs. human-written persuasive texts (three samples of human datasets combined vs. AI counterparts). Cohen's d sorted by absolute value.}
\label{tab:stylometrix_intensified_gemma_top20}
\end{table}

\begin{table}[!ht]
\centering
\scriptsize
\begin{tabular}{lcc}
\hline
\textbf{Stylometric Feature} & \textbf{Cohen's d} & \textbf{Sig.} \\
\hline
\multicolumn{3}{l}{\textit{\textbf{Gemma - Open-ended}}} \\
SY\_IMPERATIVE & 1.2484 & \checkmark \\
L\_CONT\_T & 1.2073 & \checkmark \\
L\_CONT\_A & 1.0548 & \checkmark \\
L\_FUNC\_A & -0.9226 & \checkmark \\
PS\_CAUSE & -0.8919 & \checkmark \\
VF\_INFINITIVE & 0.8048 & \checkmark \\
SENT\_D\_NP & 0.8033 & \checkmark \\
PS\_CONDITION & -0.6801 & \checkmark \\
POS\_NOUN & 0.6799 & \checkmark \\
ST\_REPETITIONS\_WORDS & -0.6654 & \checkmark \\
L\_TYPE\_TOKEN\_RATIO\_LEMMAS & 0.6102 & \checkmark \\
L\_NOUN\_PHRASES & 0.5410 & \checkmark \\
L\_LINKS & -0.5285 & \checkmark \\
L\_SINGULAR\_NOUNS & 0.5261 & \checkmark \\
SY\_INVERSE\_PATTERNS & -0.5239 & \checkmark \\
SY\_EXCLAMATION & 0.5174 & \checkmark \\
SENT\_D\_VP & 0.5048 & \checkmark \\
G\_PAST & -0.5028 & \checkmark \\
L\_HASHTAG & 0.4954 & \checkmark \\
POS\_PREP & -0.4900 & \checkmark \\
\hline
\end{tabular}
\caption{Top 20 linguistic features for AI-generated persuasive text with \textit{Open-ended} generation approach and Gemma 3 27b it model vs. human-written persuasive texts (three samples of human datasets combined vs. AI counterparts). Cohen's d sorted by absolute value.}
\label{tab:stylometrix_open_end_gemma_top20}
\end{table}

\begin{table}[!ht]
\centering
\scriptsize
\begin{tabular}{lcc}
\hline
\textbf{Stylometric Feature} & \textbf{Cohen's d} & \textbf{Sig.} \\
\hline
\multicolumn{3}{l}{\textit{\textbf{Gemini - Paraphrasing}}} \\
L\_CONT\_T & 0.8961 & \checkmark \\
L\_TYPE\_TOKEN\_RATIO\_LEMMAS & 0.8198 & \checkmark \\
L\_CONT\_A & 0.7892 & \checkmark \\
ST\_REPETITIONS\_WORDS & -0.7241 & \checkmark \\
L\_FUNC\_A & -0.6619 & \checkmark \\
L\_PUNCT\_COM & 0.5714 & \checkmark \\
SY\_INVERSE\_PATTERNS & -0.4691 & \checkmark \\
L\_ADJ\_POSITIVE & 0.4200 & \checkmark \\
POS\_ADJ & 0.3799 & \checkmark \\
L\_NOUN\_PHRASES & 0.3709 & \checkmark \\
SENT\_D\_NP & 0.3249 & \checkmark \\
POS\_PRO & -0.2846 & \checkmark \\
PS\_CONDITION & -0.2802 & \checkmark \\
L\_POSSESSIVES & 0.2743 & \checkmark \\
L\_PLURAL\_NOUNS & 0.2732 & \checkmark \\
L\_PUNCT & 0.2724 & \checkmark \\
PS\_CAUSE & -0.2699 & \checkmark \\
L\_PUNCT\_DOT & 0.2668 & \checkmark \\
VF\_INFINITIVE & -0.2396 & \checkmark \\
ASM & -0.2375 & \checkmark \\
\hline
\end{tabular}
\caption{Top 20 linguistic features for AI-generated persuasive text with \textit{Paraphrasing} generation approach and Gemini 2.0 Flash model vs. human-written persuasive texts (three samples of human datasets combined vs. AI counterparts). Cohen's d sorted by absolute value.}
\label{tab:stylometrix_paraphrased_gemini_top20}
\end{table}

\begin{table}[!ht]
\centering
\scriptsize
\begin{tabular}{lcc}
\hline
\textbf{Stylometric Feature} & \textbf{Cohen's d} & \textbf{Sig.} \\
\hline
\multicolumn{3}{l}{\textit{\textbf{Gemini - Rewriting with Subtle Persuasive Effect}}} \\
L\_CONT\_T & 1.0167 & \checkmark \\
L\_CONT\_A & 0.8716 & \checkmark \\
L\_TYPE\_TOKEN\_RATIO\_LEMMAS & 0.8497 & \checkmark \\
L\_FUNC\_A & -0.7555 & \checkmark \\
ST\_REPETITIONS\_WORDS & -0.6725 & \checkmark \\
POS\_NOUN & 0.6419 & \checkmark \\
POS\_PRO & -0.5732 & \checkmark \\
SY\_INVERSE\_PATTERNS & -0.5399 & \checkmark \\
L\_PLURAL\_NOUNS & 0.5274 & \checkmark \\
G\_ACTIVE & -0.5179 & \checkmark \\
SENT\_D\_PP & 0.5060 & \checkmark \\
L\_SECOND\_PERSON\_PRON & -0.4606 & \checkmark \\
SENT\_D\_NP & 0.4539 & \checkmark \\
CDS & -0.4498 & \checkmark \\
L\_YOU\_PRON & -0.4263 & \checkmark \\
L\_ADJ\_POSITIVE & 0.4165 & \checkmark \\
L\_PUNCT\_COM & 0.3967 & \checkmark \\
L\_NOUN\_PHRASES & 0.3906 & \checkmark \\
SY\_NARRATIVE & 0.3875 & \checkmark \\
VT\_MIGHT & 0.3785 & \checkmark \\
\hline
\end{tabular}
\caption{Top 20 linguistic features for AI-generated persuasive text with \textit{Rewriting with Subtle Persuasive Effect} generation approach and Gemini 2.0 Flash model vs. human-written persuasive texts (three samples of human datasets combined vs. AI counterparts). Cohen's d sorted by absolute value.}
\label{tab:stylometrix_rewritten_gemini_top20}
\end{table}

\begin{table}[!ht]
\centering
\scriptsize
\begin{tabular}{lcc}
\hline
\textbf{Stylometric Feature} & \textbf{Cohen's d} & \textbf{Sig.} \\
\hline
\multicolumn{3}{l}{\textit{\textbf{Gemini - Rewriting with Intensified Persuasive Effect}}} \\
L\_CONT\_T & 1.2269 & \checkmark \\
L\_CONT\_A & 1.1194 & \checkmark \\
L\_FUNC\_A & -0.9824 & \checkmark \\
L\_TYPE\_TOKEN\_RATIO\_LEMMAS & 0.8932 & \checkmark \\
L\_ADJ\_POSITIVE & 0.8740 & \checkmark \\
POS\_ADJ & 0.8168 & \checkmark \\
ST\_REPETITIONS\_WORDS & -0.7448 & \checkmark \\
L\_NOUN\_PHRASES & 0.6731 & \checkmark \\
L\_PUNCT\_COM & 0.6116 & \checkmark \\
L\_FUNC\_T & -0.5222 & \checkmark \\
POS\_PRO & -0.4680 & \checkmark \\
G\_ACTIVE & -0.4623 & \checkmark \\
PS\_CAUSE & -0.4416 & \checkmark \\
L\_ADV\_SUPERLATIVE & 0.4368 & \checkmark \\
PS\_CONDITION & -0.4344 & \checkmark \\
SY\_INVERSE\_PATTERNS & -0.4317 & \checkmark \\
SENT\_D\_NP & 0.4279 & \checkmark \\
L\_ADV\_COMPARATIVE & 0.4232 & \checkmark \\
POS\_NOUN & 0.4109 & \checkmark \\
G\_PAST & -0.4060 & \checkmark \\
\hline
\end{tabular}
\caption{Top 20 linguistic features for AI-generated persuasive text with \textit{Rewriting with Intensified Persuasive Effect} generation approach and Gemini 2.0 Flash model vs. human-written persuasive texts (three samples of human datasets combined vs. AI counterparts). Cohen's d sorted by absolute value.}
\label{tab:stylometrix_intensified_gemini_top20}
\end{table}

\begin{table}[!ht]
\centering
\scriptsize
\begin{tabular}{lcc}
\hline
\textbf{Stylometric Feature} & \textbf{Cohen's d} & \textbf{Sig.} \\
\hline
\multicolumn{3}{l}{\textit{\textbf{Gemini - Open-ended}}} \\
L\_CONT\_T & 1.3409 & \checkmark \\
L\_CONT\_A & 1.1627 & \checkmark \\
L\_TYPE\_TOKEN\_RATIO\_LEMMAS & 0.9696 & \checkmark \\
SY\_IMPERATIVE & 0.9341 & \checkmark \\
ST\_REPETITIONS\_WORDS & -0.8302 & \checkmark \\
VF\_INFINITIVE & 0.8296 & \checkmark \\
SY\_EXCLAMATION & 0.8239 & \checkmark \\
L\_FUNC\_A & -0.8082 & \checkmark \\
POS\_NOUN & 0.6966 & \checkmark \\
L\_NOUN\_PHRASES & 0.6962 & \checkmark \\
SENT\_D\_VP & 0.6846 & \checkmark \\
PS\_CAUSE & -0.6017 & \checkmark \\
PS\_CONDITION & -0.5881 & \checkmark \\
L\_SINGULAR\_NOUNS & 0.5842 & \checkmark \\
G\_ACTIVE & -0.5623 & \checkmark \\
L\_HASHTAG & 0.5397 & \checkmark \\
L\_LINKS & -0.5285 & \checkmark \\
G\_PAST & -0.5183 & \checkmark \\
SY\_INVERSE\_PATTERNS & -0.4737 & \checkmark \\
POS\_PREP & -0.4627 & \checkmark \\
\hline
\end{tabular}
\caption{Top 20 linguistic features for AI-generated persuasive text with \textit{Open-ended} generation approach and Gemini 2.0 Flash model vs. human-written persuasive texts (three samples of human datasets combined vs. AI counterparts). Cohen's d sorted by absolute value.}
\label{tab:stylometrix_open_end_gemini_top20}
\end{table}

\end{document}